\pdfoutput=1
\documentclass{article}

\usepackage[accepted]{icml2024}
\makeatletter
\renewcommand{\Notice@String}{Preprint.}
\makeatother
\usepackage[utf8]{inputenc}
\usepackage[T1]{fontenc}
\usepackage{amsmath,amssymb,amsthm,booktabs,tikz,graphicx}
\usetikzlibrary{arrows.meta,positioning,calc,fit}
\AtBeginDocument{\ifdefined\hypersetup
  \hypersetup{pdfsubject={Preprint},
    pdftitle={Recursive State-Augmented Language Models: Persistent Latent-State Adaptation of Pretrained Language Models}}\fi}

\makeatletter
\newcommand{\setv}[2]{\expandafter\gdef\csname salmv@#1\endcsname{#2}}
\newcommand{\V}[1]{\ifcsname salmv@#1\endcsname\csname salmv@#1\endcsname\else
  \PackageError{salm}{Undefined verified value `#1'}{}\fi}
\makeatother
\setv{ad/crossattention/acc}{78.1$\pm$24.7}
\setv{ad/crossattention/acc3}{88.5$\pm$4.8}
\setv{ad/crossattention/accmean}{78.1}
\setv{ad/crossattention/diff}{$-$9.7}
\setv{ad/crossattention/diffci}{[$-$40.3, +20.9]}
\setv{ad/crossattention/hours}{1.53}
\setv{ad/crossattention/kl}{0.54}
\setv{ad/crossattention/sd}{24.7}
\setv{ad/crossattention/shuf}{35.0}
\setv{ad/crossattention/shufklgap}{4.3}
\setv{ad/crossattention/worst}{34}
\setv{ad/crossattention/zero}{17.5}
\setv{ad/deepprompt/acc}{83.8$\pm$6.1}
\setv{ad/deepprompt/acc3}{87.5$\pm$0.0}
\setv{ad/deepprompt/accmean}{83.8}
\setv{ad/deepprompt/diff}{$-$4.1}
\setv{ad/deepprompt/diffci}{[$-$11.4, +3.3]}
\setv{ad/deepprompt/hours}{2.10}
\setv{ad/deepprompt/kl}{0.43}
\setv{ad/deepprompt/sd}{6.1}
\setv{ad/deepprompt/shuf}{30.9}
\setv{ad/deepprompt/shufklgap}{3.3}
\setv{ad/deepprompt/worst}{73}
\setv{ad/deepprompt/zero}{31.2}
\setv{ad/film/acc}{46.6$\pm$19.3}
\setv{ad/film/acc3}{51.6$\pm$25.1}
\setv{ad/film/accmean}{46.6}
\setv{ad/film/diff}{$-$41.2}
\setv{ad/film/diffci}{[$-$64.4, $-$18.1]}
\setv{ad/film/hours}{1.56}
\setv{ad/film/kl}{1.11}
\setv{ad/film/sd}{19.3}
\setv{ad/film/shuf}{31.9}
\setv{ad/film/shufklgap}{0.7}
\setv{ad/film/worst}{31}
\setv{ad/film/zero}{31.2}
\setv{ad/fullcontextlora/acc}{87.8$\pm$0.7}
\setv{ad/fullcontextlora/acc10}{89.2$\pm$1.7}
\setv{ad/fullcontextlora/acc3}{88.0$\pm$0.9}
\setv{ad/fullcontextlora/accmean}{87.8}
\setv{ad/fullcontextlora/hours}{0.32}
\setv{ad/fullcontextlora/kl}{0.42}
\setv{ad/fullcontextlora/sd}{0.7}
\setv{ad/fullcontextlora/shuf}{34.7}
\setv{ad/fullcontextlora/shufklgap}{6.8}
\setv{ad/fullcontextlora/worst}{88}
\setv{ad/fullcontextlora/zero}{25.6}
\setv{ad/gatedlora/acc}{88.4$\pm$3.2}
\setv{ad/gatedlora/acc10}{87.7$\pm$3.3}
\setv{ad/gatedlora/acc3}{89.1$\pm$4.1}
\setv{ad/gatedlora/accmean}{88.4}
\setv{ad/gatedlora/diff}{+0.6}
\setv{ad/gatedlora/diffci}{[$-$2.9, +4.1]}
\setv{ad/gatedlora/hours}{0.91}
\setv{ad/gatedlora/kl}{0.38}
\setv{ad/gatedlora/sd}{3.2}
\setv{ad/gatedlora/shuf}{34.1}
\setv{ad/gatedlora/shufklgap}{5.5}
\setv{ad/gatedlora/worst}{84}
\setv{ad/gatedlora/zero}{31.2}
\setv{ad/generatedlora/acc}{74.7$\pm$15.0}
\setv{ad/generatedlora/acc3}{79.2$\pm$7.0}
\setv{ad/generatedlora/accmean}{74.7}
\setv{ad/generatedlora/diff}{$-$13.1}
\setv{ad/generatedlora/diffci}{[$-$31.8, +5.6]}
\setv{ad/generatedlora/hours}{0.71}
\setv{ad/generatedlora/kl}{0.75}
\setv{ad/generatedlora/sd}{15.0}
\setv{ad/generatedlora/shuf}{30.6}
\setv{ad/generatedlora/shufklgap}{4.5}
\setv{ad/generatedlora/worst}{50}
\setv{ad/generatedlora/zero}{25.6}
\setv{ad/majority}{34.4}
\setv{ad/n}{64}
\setv{ad/prefix/acc}{55.6$\pm$16.1}
\setv{ad/prefix/acc3}{62.5$\pm$14.9}
\setv{ad/prefix/accmean}{55.6}
\setv{ad/prefix/diff}{$-$32.2}
\setv{ad/prefix/diffci}{[$-$52.2, $-$12.1]}
\setv{ad/prefix/hours}{2.61}
\setv{ad/prefix/kl}{0.91}
\setv{ad/prefix/sd}{16.1}
\setv{ad/prefix/shuf}{27.8}
\setv{ad/prefix/shufklgap}{1.9}
\setv{ad/prefix/worst}{34}
\setv{ad/prefix/zero}{31.2}
\setv{ad/softprompt/acc}{82.5$\pm$9.1}
\setv{ad/softprompt/acc3}{87.5$\pm$5.6}
\setv{ad/softprompt/accmean}{82.5}
\setv{ad/softprompt/diff}{$-$5.3}
\setv{ad/softprompt/diffci}{[$-$16.3, +5.6]}
\setv{ad/softprompt/hours}{0.90}
\setv{ad/softprompt/kl}{0.65}
\setv{ad/softprompt/sd}{9.1}
\setv{ad/softprompt/shuf}{36.2}
\setv{ad/softprompt/shufklgap}{6.7}
\setv{ad/softprompt/worst}{69}
\setv{ad/softprompt/zero}{31.2}
\setv{ad/testcounts}{22, 20, 15, 7}
\setv{ad/zeroconst}{31.2}
\setv{cal/det/maxabs}{0.07}
\setv{cal/det/over}{0.07}
\setv{cal/det/under}{0.04}
\setv{cal/sto/maxabs}{0.04}
\setv{cal/sto/over}{0.04}
\setv{cal/sto/under}{0.02}
\setv{cmp/kvmax}{12.1}
\setv{cmp/kvmin}{3.1}
\setv{cmp/peakmax}{1.05}
\setv{cmp/peakmin}{1.04}
\setv{cmp/ratiomax}{8.2}
\setv{cmp/ratiomin}{6.4}
\setv{cmp/statekb}{3.3}
\setv{lh/B4/diff}{$-$6.6}
\setv{lh/B4/diffci}{[$-$8.4, $-$4.9]}
\setv{lh/B4/full}{89.0$\pm$2.2}
\setv{lh/B4/gated}{82.3$\pm$2.1}
\setv{lh/B4/gatedzero}{13.3}
\setv{lh/B6/diff}{$-$22.3}
\setv{lh/B6/diffci}{[$-$28.1, $-$16.6]}
\setv{lh/B6/full}{87.2$\pm$2.0}
\setv{lh/B6/gated}{64.8$\pm$8.6}
\setv{lh/B6/gatedzero}{7.0}
\setv{lh/B8/diff}{$-$22.1}
\setv{lh/B8/diffci}{[$-$26.4, $-$17.8]}
\setv{lh/B8/full}{84.3$\pm$3.6}
\setv{lh/B8/gated}{62.2$\pm$6.4}
\setv{lh/B8/gatedzero}{9.4}
\setv{lh/n}{128}
\setv{lh/zero/max}{13.3}
\setv{lh/zero/min}{7.0}
\setv{lh3/A12/full}{83.6$\pm$2.3}
\setv{lh3/A12/gated}{78.9$\pm$4.3}
\setv{lh3/B12/full}{84.6$\pm$11.1}
\setv{lh3/B12/gated}{55.7$\pm$4.3}
\setv{lt/diff/acc}{$-$4.0}
\setv{lt/diff/accabs}{4.0}
\setv{lt/diff/accci}{[$-$5.1, $-$2.9]}
\setv{lt/diff/acchb}{[$-$5.9, $-$2.2]}
\setv{lt/diff/kl}{+0.13}
\setv{lt/diff/klci}{[+0.08, +0.18]}
\setv{lt/full/acc}{94.4$\pm$1.2}
\setv{lt/full/kl}{0.22}
\setv{lt/full/shuf}{24.6$\pm$0.6}
\setv{lt/full/zero}{24.5$\pm$1.7}
\setv{lt/gated/acc}{90.4$\pm$2.2}
\setv{lt/gated/kl}{0.35}
\setv{lt/gated/maxshuf}{25.0}
\setv{lt/gated/maxzero}{24.6}
\setv{lt/gated/shuf}{24.6$\pm$0.5}
\setv{lt/gated/zero}{24.6$\pm$0.0}
\setv{lt/n}{512}
\setv{lt/pairedp}{$<$0.001}
\setv{mrc/diff}{0.050}
\setv{mrc/diffci}{[0.030, 0.071]}
\setv{mrc/fc/f1}{61.6}
\setv{mrc/fc/nll}{0.592}
\setv{mrc/fc/nopassnll}{0.729}
\setv{mrc/fc/shufnll}{0.898}
\setv{mrc/npairs}{4{,}848}
\setv{mrc/npassages}{83}
\setv{mrc/nseeds}{4}
\setv{mrc/nval}{259}
\setv{mrc/salm/f1}{53.2}
\setv{mrc/salm/nll}{0.642}
\setv{mrc/salm/zeronll}{1.016}
\setv{mrc/shuf}{0.002}
\setv{mrc/shufci}{[$-$0.002, +0.007]}
\setv{of/A/acc}{45.6}
\setv{of/A/accmax}{49.2}
\setv{of/A/accmin}{38.9}
\setv{of/A/accsd}{5.8}
\setv{of/A/best}{8, 15, 11}
\setv{of/A/distinct_particle_maps}{3.34}
\setv{of/A/distinct_particle_mapsmax}{3.54}
\setv{of/A/distinct_particle_mapsmin}{2.92}
\setv{of/A/distinct_particle_mapssd}{0.36}
\setv{of/A/ece}{0.085}
\setv{of/A/ecemax}{0.101}
\setv{of/A/ecemin}{0.072}
\setv{of/A/ecesd}{0.014}
\setv{of/A/epochs}{14, 21, 17}
\setv{of/A/ess}{1.90}
\setv{of/A/essmax}{2.00}
\setv{of/A/essmin}{1.77}
\setv{of/A/esssd}{0.12}
\setv{of/A/fracremoved}{47}
\setv{of/A/kl}{0.246}
\setv{of/A/klmax}{0.303}
\setv{of/A/klmin}{0.191}
\setv{of/A/klsd}{0.056}
\setv{of/A/nll}{1.150}
\setv{of/A/nllmax}{1.206}
\setv{of/A/nllmin}{1.099}
\setv{of/A/nllsd}{0.054}
\setv{of/A/particle_effective_codes}{2.18}
\setv{of/A/particle_effective_codesmax}{2.50}
\setv{of/A/particle_effective_codesmin}{1.96}
\setv{of/A/particle_effective_codessd}{0.29}
\setv{of/A/particle_max_prob}{0.67}
\setv{of/A/particle_max_probmax}{0.73}
\setv{of/A/particle_max_probmin}{0.55}
\setv{of/A/particle_max_probsd}{0.10}
\setv{of/A/shuffle_gap}{0.75}
\setv{of/A/shuffle_gapmax}{0.90}
\setv{of/A/shuffle_gapmin}{0.60}
\setv{of/A/shuffle_gapsd}{0.15}
\setv{of/B/acc}{49.0}
\setv{of/B/accmax}{50.6}
\setv{of/B/accmin}{47.9}
\setv{of/B/accsd}{1.4}
\setv{of/B/best}{12, 12, 11}
\setv{of/B/distinct_particle_maps}{3.82}
\setv{of/B/distinct_particle_mapsmax}{3.98}
\setv{of/B/distinct_particle_mapsmin}{3.66}
\setv{of/B/distinct_particle_mapssd}{0.16}
\setv{of/B/ece}{0.080}
\setv{of/B/ecemax}{0.081}
\setv{of/B/ecemin}{0.079}
\setv{of/B/ecesd}{0.001}
\setv{of/B/epochs}{18, 18, 17}
\setv{of/B/ess}{2.01}
\setv{of/B/essmax}{2.03}
\setv{of/B/essmin}{1.99}
\setv{of/B/esssd}{0.02}
\setv{of/B/fracremoved}{58}
\setv{of/B/kl}{0.197}
\setv{of/B/klmax}{0.214}
\setv{of/B/klmin}{0.174}
\setv{of/B/klsd}{0.021}
\setv{of/B/nll}{1.109}
\setv{of/B/nllmax}{1.125}
\setv{of/B/nllmin}{1.089}
\setv{of/B/nllsd}{0.018}
\setv{of/B/particle_effective_codes}{1.80}
\setv{of/B/particle_effective_codesmax}{1.82}
\setv{of/B/particle_effective_codesmin}{1.78}
\setv{of/B/particle_effective_codessd}{0.02}
\setv{of/B/particle_max_prob}{0.75}
\setv{of/B/particle_max_probmax}{0.77}
\setv{of/B/particle_max_probmin}{0.73}
\setv{of/B/particle_max_probsd}{0.02}
\setv{of/B/shuffle_gap}{0.88}
\setv{of/B/shuffle_gapmax}{0.88}
\setv{of/B/shuffle_gapmin}{0.87}
\setv{of/B/shuffle_gapsd}{0.01}
\setv{of/C/acc}{52.2}
\setv{of/C/accmax}{53.1}
\setv{of/C/accmin}{51.2}
\setv{of/C/accsd}{1.0}
\setv{of/C/best}{12, 11, 8}
\setv{of/C/distinct_particle_maps}{3.89}
\setv{of/C/distinct_particle_mapsmax}{4.00}
\setv{of/C/distinct_particle_mapsmin}{3.77}
\setv{of/C/distinct_particle_mapssd}{0.12}
\setv{of/C/ece}{0.058}
\setv{of/C/ecemax}{0.066}
\setv{of/C/ecemin}{0.047}
\setv{of/C/ecesd}{0.010}
\setv{of/C/epochs}{18, 17, 14}
\setv{of/C/ess}{2.14}
\setv{of/C/essmax}{2.23}
\setv{of/C/essmin}{2.09}
\setv{of/C/esssd}{0.08}
\setv{of/C/fracremoved}{77}
\setv{of/C/kl}{0.108}
\setv{of/C/klmax}{0.117}
\setv{of/C/klmin}{0.093}
\setv{of/C/klsd}{0.013}
\setv{of/C/nll}{1.011}
\setv{of/C/nllmax}{1.029}
\setv{of/C/nllmin}{0.992}
\setv{of/C/nllsd}{0.019}
\setv{of/C/particle_effective_codes}{1.67}
\setv{of/C/particle_effective_codesmax}{1.74}
\setv{of/C/particle_effective_codesmin}{1.56}
\setv{of/C/particle_effective_codessd}{0.10}
\setv{of/C/particle_max_prob}{0.79}
\setv{of/C/particle_max_probmax}{0.84}
\setv{of/C/particle_max_probmin}{0.76}
\setv{of/C/particle_max_probsd}{0.04}
\setv{of/C/shuffle_gap}{0.99}
\setv{of/C/shuffle_gapmax}{1.10}
\setv{of/C/shuffle_gapmin}{0.89}
\setv{of/C/shuffle_gapsd}{0.10}
\setv{of/M4K8/acc}{47.9}
\setv{of/M4K8/accmax}{50.4}
\setv{of/M4K8/accmin}{46.1}
\setv{of/M4K8/accsd}{2.2}
\setv{of/M4K8/best}{13, 11, 8}
\setv{of/M4K8/distinct_particle_maps}{3.55}
\setv{of/M4K8/distinct_particle_mapsmax}{3.95}
\setv{of/M4K8/distinct_particle_mapsmin}{3.28}
\setv{of/M4K8/distinct_particle_mapssd}{0.35}
\setv{of/M4K8/ece}{0.093}
\setv{of/M4K8/ecemax}{0.103}
\setv{of/M4K8/ecemin}{0.077}
\setv{of/M4K8/ecesd}{0.014}
\setv{of/M4K8/epochs}{19, 17, 14}
\setv{of/M4K8/ess}{4.13}
\setv{of/M4K8/essmax}{4.16}
\setv{of/M4K8/essmin}{4.09}
\setv{of/M4K8/esssd}{0.04}
\setv{of/M4K8/fracremoved}{60}
\setv{of/M4K8/kl}{0.185}
\setv{of/M4K8/klmax}{0.216}
\setv{of/M4K8/klmin}{0.168}
\setv{of/M4K8/klsd}{0.026}
\setv{of/M4K8/nll}{1.096}
\setv{of/M4K8/nllmax}{1.123}
\setv{of/M4K8/nllmin}{1.072}
\setv{of/M4K8/nllsd}{0.026}
\setv{of/M4K8/particle_effective_codes}{1.79}
\setv{of/M4K8/particle_effective_codesmax}{1.83}
\setv{of/M4K8/particle_effective_codesmin}{1.75}
\setv{of/M4K8/particle_effective_codessd}{0.04}
\setv{of/M4K8/particle_max_prob}{0.76}
\setv{of/M4K8/particle_max_probmax}{0.76}
\setv{of/M4K8/particle_max_probmin}{0.75}
\setv{of/M4K8/particle_max_probsd}{0.00}
\setv{of/M4K8/shuffle_gap}{0.87}
\setv{of/M4K8/shuffle_gapmax}{0.89}
\setv{of/M4K8/shuffle_gapmin}{0.86}
\setv{of/M4K8/shuffle_gapsd}{0.02}
\setv{of/M8K4/acc}{27.0}
\setv{of/M8K4/accmax}{27.1}
\setv{of/M8K4/accmin}{26.6}
\setv{of/M8K4/accsd}{0.3}
\setv{of/M8K4/best}{7, 13, 11}
\setv{of/M8K4/distinct_particle_maps}{3.76}
\setv{of/M8K4/distinct_particle_mapsmax}{3.96}
\setv{of/M8K4/distinct_particle_mapsmin}{3.38}
\setv{of/M8K4/distinct_particle_mapssd}{0.33}
\setv{of/M8K4/ece}{0.082}
\setv{of/M8K4/ecemax}{0.100}
\setv{of/M8K4/ecemin}{0.073}
\setv{of/M8K4/ecesd}{0.015}
\setv{of/M8K4/epochs}{13, 17, 17}
\setv{of/M8K4/ess}{1.87}
\setv{of/M8K4/essmax}{1.94}
\setv{of/M8K4/essmin}{1.76}
\setv{of/M8K4/esssd}{0.10}
\setv{of/M8K4/fracremoved}{44}
\setv{of/M8K4/kl}{0.383}
\setv{of/M8K4/klmax}{0.459}
\setv{of/M8K4/klmin}{0.337}
\setv{of/M8K4/klsd}{0.067}
\setv{of/M8K4/nll}{1.766}
\setv{of/M8K4/nllmax}{1.819}
\setv{of/M8K4/nllmin}{1.733}
\setv{of/M8K4/nllsd}{0.046}
\setv{of/M8K4/particle_effective_codes}{3.87}
\setv{of/M8K4/particle_effective_codesmax}{4.18}
\setv{of/M8K4/particle_effective_codesmin}{3.39}
\setv{of/M8K4/particle_effective_codessd}{0.42}
\setv{of/M8K4/particle_max_prob}{0.46}
\setv{of/M8K4/particle_max_probmax}{0.48}
\setv{of/M8K4/particle_max_probmin}{0.45}
\setv{of/M8K4/particle_max_probsd}{0.02}
\setv{of/M8K4/shuffle_gap}{0.91}
\setv{of/M8K4/shuffle_gapmax}{1.00}
\setv{of/M8K4/shuffle_gapmin}{0.83}
\setv{of/M8K4/shuffle_gapsd}{0.09}
\setv{of/bayes4/acc}{57.8}
\setv{of/bayes4/nll}{0.904}
\setv{of/bayes8/acc}{46.9}
\setv{of/bayes8/nll}{1.380}
\setv{of/det/acc}{26.3}
\setv{of/det/accmax}{27.1}
\setv{of/det/accmin}{24.8}
\setv{of/det/accsd}{1.3}
\setv{of/det/best}{3, 8, 2}
\setv{of/det/ece}{0.019}
\setv{of/det/ecemax}{0.027}
\setv{of/det/ecemin}{0.012}
\setv{of/det/ecesd}{0.008}
\setv{of/det/epochs}{10, 14, 10}
\setv{of/det/ess}{1.00}
\setv{of/det/essmax}{1.00}
\setv{of/det/essmin}{1.00}
\setv{of/det/esssd}{0.00}
\setv{of/det/kl}{0.475}
\setv{of/det/klmax}{0.477}
\setv{of/det/klmin}{0.472}
\setv{of/det/klsd}{0.003}
\setv{of/det/nll}{1.388}
\setv{of/det/nllmax}{1.391}
\setv{of/det/nllmin}{1.386}
\setv{of/det/nllsd}{0.003}
\setv{of/det/shuffle_gap}{0.00}
\setv{of/det/shuffle_gapmax}{0.00}
\setv{of/det/shuffle_gapmin}{0.00}
\setv{of/det/shuffle_gapsd}{0.00}
\setv{of/diff/B-A/acc}{+3.5}
\setv{of/diff/B-A/accci}{[$-$10.0, +16.9]}
\setv{of/diff/B-A/distinct_particle_maps}{+0.48}
\setv{of/diff/B-A/distinct_particle_mapsci}{[$-$0.28, +1.24]}
\setv{of/diff/B-A/kl}{$-$0.049}
\setv{of/diff/B-A/klci}{[$-$0.171, +0.073]}
\setv{of/diff/B-A/nll}{$-$0.041}
\setv{of/diff/B-A/nllci}{[$-$0.160, +0.078]}
\setv{of/diff/C-A/acc}{+6.6}
\setv{of/diff/C-A/accci}{[$-$7.3, +20.6]}
\setv{of/diff/C-A/distinct_particle_maps}{+0.55}
\setv{of/diff/C-A/distinct_particle_mapsci}{[$-$0.24, +1.34]}
\setv{of/diff/C-A/kl}{$-$0.138}
\setv{of/diff/C-A/klci}{[$-$0.269, $-$0.008]}
\setv{of/diff/C-A/nll}{$-$0.139}
\setv{of/diff/C-A/nllci}{[$-$0.257, $-$0.020]}
\setv{of/diff/C-B/acc}{+3.2}
\setv{of/diff/C-B/accci}{[+0.3, +6.1]}
\setv{of/diff/C-B/distinct_particle_maps}{+0.07}
\setv{of/diff/C-B/distinct_particle_mapsci}{[$-$0.25, +0.39]}
\setv{of/diff/C-B/kl}{$-$0.089}
\setv{of/diff/C-B/klci}{[$-$0.132, $-$0.047]}
\setv{of/diff/C-B/nll}{$-$0.098}
\setv{of/diff/C-B/nllci}{[$-$0.139, $-$0.056]}
\setv{of/diff/M4K8-B/acc}{$-$1.1}
\setv{of/diff/M4K8-B/accci}{[$-$5.6, +3.4]}
\setv{of/diff/M4K8-B/distinct_particle_maps}{$-$0.26}
\setv{of/diff/M4K8-B/distinct_particle_mapsci}{[$-$1.00, +0.48]}
\setv{of/diff/M4K8-B/kl}{$-$0.012}
\setv{of/diff/M4K8-B/klci}{[$-$0.067, +0.043]}
\setv{of/diff/M4K8-B/nll}{$-$0.014}
\setv{of/diff/M4K8-B/nllci}{[$-$0.066, +0.039]}
\setv{of/tied4}{50.4}
\setv{of/tied8}{64.8}
\setv{of/unif4/kl}{0.469}
\setv{of/unif8/kl}{0.684}
\setv{ret/d1min}{0.84}
\setv{ret/episodes}{512}
\setv{ret/h4/cm1}{1.00}
\setv{ret/h4/cmT}{0.97}
\setv{ret/h5/cm1}{0.96}
\setv{ret/h5/cmT}{0.73}
\setv{ret/h5/d1T}{0.82}
\setv{ret/h5/d1min}{0.72}
\setv{ret/hybrid/max}{0.98}
\setv{ret/hybrid/min}{0.74}
\setv{ret/late/max}{0.78}
\setv{ret/late/min}{0.60}
\setv{ret/reset/max}{44}
\setv{ret/reset/min}{35}
\setv{ret/swap/max}{31}
\setv{ret/swap/min}{20}
\setv{sto/bayes/acc}{55.5}
\setv{sto/bayes/accu}{68.4}
\setv{sto/bayes/brier}{0.534}
\setv{sto/bayes/brieru}{0.439}
\setv{sto/bayes/ece}{0.037}
\setv{sto/bayes/eceu}{0.044}
\setv{sto/bayes/nll}{0.970}
\setv{sto/bayes/nllu}{0.812}
\setv{sto/det/acc}{55.4}
\setv{sto/det/accsd}{0.6}
\setv{sto/det/accu}{67.9}
\setv{sto/det/accusd}{0.8}
\setv{sto/det/best}{12, 21, 21, 24, 14}
\setv{sto/det/brier}{0.555}
\setv{sto/det/briersd}{0.005}
\setv{sto/det/brieru}{0.465}
\setv{sto/det/brierusd}{0.005}
\setv{sto/det/ece}{0.048}
\setv{sto/det/ecesd}{0.012}
\setv{sto/det/eceu}{0.068}
\setv{sto/det/eceusd}{0.026}
\setv{sto/det/entropy}{0.947}
\setv{sto/det/entropysd}{0.026}
\setv{sto/det/entropyu}{0.807}
\setv{sto/det/entropyusd}{0.037}
\setv{sto/det/epochs}{18, 24, 27, 26, 20}
\setv{sto/det/flippredmax}{1.49}
\setv{sto/det/flippredmin}{1.38}
\setv{sto/det/flipstatemax}{2.49}
\setv{sto/det/flipstatemin}{2.02}
\setv{sto/det/fracremoved}{90}
\setv{sto/det/gain}{0.00}
\setv{sto/det/hours}{2.6}
\setv{sto/det/infer}{0.19}
\setv{sto/det/kl}{0.042}
\setv{sto/det/klsd}{0.007}
\setv{sto/det/klu}{0.033}
\setv{sto/det/klusd}{0.007}
\setv{sto/det/mapagree}{99.8}
\setv{sto/det/mapagreemin}{99.6}
\setv{sto/det/mapagreeminu}{98.8}
\setv{sto/det/mapagreeu}{99.8}
\setv{sto/det/nll}{1.011}
\setv{sto/det/nllsd}{0.009}
\setv{sto/det/nllu}{0.861}
\setv{sto/det/nllusd}{0.010}
\setv{sto/det/params}{2{,}223{,}012}
\setv{sto/det/post_cal_err}{0.081}
\setv{sto/det/post_cal_errsd}{0.010}
\setv{sto/det/post_cal_erru}{0.072}
\setv{sto/det/post_cal_errusd}{0.007}
\setv{sto/det/shuf_kl}{1.055}
\setv{sto/det/shuf_klsd}{0.056}
\setv{sto/det/shuf_klu}{1.317}
\setv{sto/det/shuf_klusd}{0.092}
\setv{sto/det/zero_kl}{0.573}
\setv{sto/det/zero_klsd}{0.105}
\setv{sto/det/zero_klu}{0.720}
\setv{sto/det/zero_klusd}{0.088}
\setv{sto/diff/acc}{$-$1.6}
\setv{sto/diff/accabs}{1.6}
\setv{sto/diff/accci}{[$-$2.9, $-$0.3]}
\setv{sto/diff/accu}{+2.2}
\setv{sto/diff/accuabs}{2.2}
\setv{sto/diff/accuci}{[+1.4, +3.0]}
\setv{sto/diff/brier}{$-$0.015}
\setv{sto/diff/brierci}{[$-$0.020, $-$0.010]}
\setv{sto/diff/ece}{+0.015}
\setv{sto/diff/ececi}{[$-$0.015, +0.045]}
\setv{sto/diff/kl}{$-$0.032}
\setv{sto/diff/klci}{[$-$0.039, $-$0.025]}
\setv{sto/diff/klu}{$-$0.024}
\setv{sto/diff/kluci}{[$-$0.033, $-$0.015]}
\setv{sto/diff/nll}{$-$0.030}
\setv{sto/diff/nllci}{[$-$0.041, $-$0.020]}
\setv{sto/diff/post_cal_err}{$-$0.045}
\setv{sto/diff/post_cal_errci}{[$-$0.058, $-$0.033]}
\setv{sto/diff/trkl}{+0.001}
\setv{sto/diff/trklci}{[$-$0.004, +0.007]}
\setv{sto/flip/oracle}{1.45}
\setv{sto/inferratio}{3.1}
\setv{sto/initdist}{3.4}
\setv{sto/n}{512}
\setv{sto/nu}{256}
\setv{sto/postent}{0.98}
\setv{sto/postentu}{0.83}
\setv{sto/sto/acc}{53.9}
\setv{sto/sto/accsd}{0.8}
\setv{sto/sto/accu}{70.1}
\setv{sto/sto/accusd}{1.0}
\setv{sto/sto/best}{10, 8, 10, 10, 14}
\setv{sto/sto/brier}{0.540}
\setv{sto/sto/briersd}{0.001}
\setv{sto/sto/brieru}{0.441}
\setv{sto/sto/brierusd}{0.002}
\setv{sto/sto/cohmax}{0.99}
\setv{sto/sto/cohmin}{0.98}
\setv{sto/sto/disp}{0.29}
\setv{sto/sto/distinctmax}{4.00}
\setv{sto/sto/distinctmin}{4.00}
\setv{sto/sto/ece}{0.063}
\setv{sto/sto/ecesd}{0.018}
\setv{sto/sto/eceu}{0.058}
\setv{sto/sto/eceusd}{0.015}
\setv{sto/sto/entropy}{0.958}
\setv{sto/sto/entropysd}{0.025}
\setv{sto/sto/entropyu}{0.808}
\setv{sto/sto/entropyusd}{0.035}
\setv{sto/sto/epochs}{13, 14, 11, 16, 16}
\setv{sto/sto/flippredmax}{1.52}
\setv{sto/sto/flippredmin}{1.43}
\setv{sto/sto/flipstatemax}{0.56}
\setv{sto/sto/flipstatemin}{0.04}
\setv{sto/sto/fracremoved}{98}
\setv{sto/sto/gain}{1.89}
\setv{sto/sto/hours}{4.1}
\setv{sto/sto/infer}{0.60}
\setv{sto/sto/kl}{0.009}
\setv{sto/sto/klK1}{5.33}
\setv{sto/sto/klK8}{0.008}
\setv{sto/sto/klsd}{0.002}
\setv{sto/sto/klu}{0.009}
\setv{sto/sto/klusd}{0.002}
\setv{sto/sto/mapagree}{100.0}
\setv{sto/sto/mapagreemin}{100.0}
\setv{sto/sto/mapagreeminu}{100.0}
\setv{sto/sto/mapagreeu}{100.0}
\setv{sto/sto/nll}{0.980}
\setv{sto/sto/nllsd}{0.002}
\setv{sto/sto/nllu}{0.823}
\setv{sto/sto/nllusd}{0.007}
\setv{sto/sto/params}{2{,}223{,}284}
\setv{sto/sto/pmaxmin}{0.97}
\setv{sto/sto/post_cal_err}{0.035}
\setv{sto/sto/post_cal_errsd}{0.006}
\setv{sto/sto/post_cal_erru}{0.030}
\setv{sto/sto/post_cal_errusd}{0.006}
\setv{sto/sto/shuf_kl}{3.992}
\setv{sto/sto/shuf_klsd}{0.559}
\setv{sto/sto/shuf_klu}{4.412}
\setv{sto/sto/shuf_klusd}{0.564}
\setv{sto/sto/zero_kl}{1.317}
\setv{sto/sto/zero_klsd}{0.266}
\setv{sto/sto/zero_klu}{1.834}
\setv{sto/sto/zero_klusd}{0.281}
\setv{sto/tied}{53.1}
\setv{sto/tiedu}{18.0}
\setv{sto/tr/acc}{54.6}
\setv{sto/tr/accsd}{0.4}
\setv{sto/tr/accu}{69.7}
\setv{sto/tr/accusd}{1.2}
\setv{sto/tr/best}{10, 8, 9}
\setv{sto/tr/brier}{0.541}
\setv{sto/tr/briersd}{0.001}
\setv{sto/tr/brieru}{0.440}
\setv{sto/tr/brierusd}{0.004}
\setv{sto/tr/cohmax}{0.99}
\setv{sto/tr/cohmin}{0.99}
\setv{sto/tr/disp}{0.30}
\setv{sto/tr/ece}{0.054}
\setv{sto/tr/ecesd}{0.005}
\setv{sto/tr/eceu}{0.049}
\setv{sto/tr/eceusd}{0.019}
\setv{sto/tr/entropy}{0.983}
\setv{sto/tr/entropysd}{0.007}
\setv{sto/tr/entropyu}{0.842}
\setv{sto/tr/entropyusd}{0.014}
\setv{sto/tr/epochs}{16, 14, 15}
\setv{sto/tr/flippredmax}{1.46}
\setv{sto/tr/flippredmin}{1.43}
\setv{sto/tr/flipstatemax}{0.15}
\setv{sto/tr/flipstatemin}{0.00}
\setv{sto/tr/gain}{1.17}
\setv{sto/tr/hours}{5.0}
\setv{sto/tr/infer}{0.60}
\setv{sto/tr/kl}{0.010}
\setv{sto/tr/klsd}{0.002}
\setv{sto/tr/klu}{0.009}
\setv{sto/tr/klusd}{0.002}
\setv{sto/tr/nll}{0.980}
\setv{sto/tr/nllsd}{0.000}
\setv{sto/tr/nllu}{0.822}
\setv{sto/tr/nllusd}{0.007}
\setv{sto/tr/params}{2{,}223{,}284}
\setv{sto/tr/post_cal_err}{0.032}
\setv{sto/tr/post_cal_errsd}{0.001}
\setv{sto/tr/post_cal_erru}{0.026}
\setv{sto/tr/post_cal_errusd}{0.005}
\setv{sto/tr/shuf_kl}{2.095}
\setv{sto/tr/shuf_klsd}{0.315}
\setv{sto/tr/shuf_klu}{2.218}
\setv{sto/tr/shuf_klusd}{0.161}
\setv{sto/tr/zero_kl}{0.913}
\setv{sto/tr/zero_klsd}{0.145}
\setv{sto/tr/zero_klu}{1.190}
\setv{sto/tr/zero_klusd}{0.189}
\setv{sto/trainratio}{1.5}
\setv{sto/unif/kl}{0.411}
\setv{sto/unif/klu}{0.556}
\setv{v3/early/fullmax}{46}
\setv{v3/early/fullmin}{25}
\setv{v3/early/salmmax}{45}
\setv{v3/early/salmmin}{32}
\setv{v3/h1/diff}{$-$4.4}
\setv{v3/h1/diffabs}{4.4}
\setv{v3/h1/diffci}{[$-$9.1, +0.4]}
\setv{v3/h1/full}{96.4$\pm$0.7}
\setv{v3/h1/fullearly}{33.7}
\setv{v3/h1/fullhours}{0.08}
\setv{v3/h1/fullmean}{96.4}
\setv{v3/h1/salm}{92.0$\pm$3.8}
\setv{v3/h1/salmearly}{43.4}
\setv{v3/h1/salmhours}{0.27}
\setv{v3/h1/salmmean}{92.0}
\setv{v3/h1/shufpen}{24.4}
\setv{v3/h2/diff}{$-$4.3}
\setv{v3/h2/diffabs}{4.3}
\setv{v3/h2/diffci}{[$-$8.3, $-$0.2]}
\setv{v3/h2/full}{93.8$\pm$0.3}
\setv{v3/h2/fullearly}{35.4}
\setv{v3/h2/fullhours}{0.14}
\setv{v3/h2/fullmean}{93.8}
\setv{v3/h2/salm}{89.6$\pm$3.4}
\setv{v3/h2/salmearly}{42.7}
\setv{v3/h2/salmhours}{0.72}
\setv{v3/h2/salmmean}{89.6}
\setv{v3/h2/shufpen}{23.5}
\setv{v3/h3/diff}{$-$4.0}
\setv{v3/h3/diffabs}{4.0}
\setv{v3/h3/diffci}{[$-$8.5, +0.5]}
\setv{v3/h3/full}{95.2$\pm$1.5}
\setv{v3/h3/fullearly}{29.8}
\setv{v3/h3/fullhours}{0.19}
\setv{v3/h3/fullmean}{95.2}
\setv{v3/h3/salm}{91.2$\pm$3.7}
\setv{v3/h3/salmearly}{45.4}
\setv{v3/h3/salmhours}{1.30}
\setv{v3/h3/salmmean}{91.2}
\setv{v3/h3/shufpen}{23.2}
\setv{v3/h4/diff}{+0.7}
\setv{v3/h4/diffabs}{0.7}
\setv{v3/h4/diffci}{[$-$0.2, +1.6]}
\setv{v3/h4/full}{96.4$\pm$0.7}
\setv{v3/h4/fullearly}{46.4}
\setv{v3/h4/fullhours}{0.10}
\setv{v3/h4/fullmean}{96.4}
\setv{v3/h4/salm}{97.2$\pm$1.2}
\setv{v3/h4/salmearly}{36.4}
\setv{v3/h4/salmhours}{0.61}
\setv{v3/h4/salmmean}{97.2}
\setv{v3/h4/shufpen}{21.9}
\setv{v3/h5/diff}{$-$20.9}
\setv{v3/h5/diffabs}{20.9}
\setv{v3/h5/diffci}{[$-$24.1, $-$17.6]}
\setv{v3/h5/full}{94.2$\pm$0.5}
\setv{v3/h5/fullearly}{24.7}
\setv{v3/h5/fullhours}{0.14}
\setv{v3/h5/fullmean}{94.2}
\setv{v3/h5/salm}{73.4$\pm$2.4}
\setv{v3/h5/salmearly}{31.8}
\setv{v3/h5/salmhours}{0.75}
\setv{v3/h5/salmmean}{73.4}
\setv{v3/h5/shufpen}{20.4}
\setv{v3/h5x/best}{27}
\setv{v3/h5x/last}{35}
\setv{v3/n}{512}
\setv{v3/shufpen/max}{24}
\setv{v3/shufpen/min}{20}

\newtheorem{assumption}{Assumption}
\newtheorem{proposition}{Proposition}

\theoremstyle{remark}
\newtheorem{remark}{Remark}

\newcommand{\E}{\mathbb{E}}
\newcommand{\KL}{\mathrm{KL}}
\newcommand{\pstar}{p^{\star}}

\icmltitlerunning{Recursive State-Augmented Language Models}

\begin{document}

\twocolumn[
\icmltitle{
Recursive State-Augmented Language Models:\\
Persistent Latent-State Adaptation of Pretrained Language Models
}

\begin{icmlauthorlist}
\icmlauthor{Omer Haq}{ind}
\end{icmlauthorlist}

\icmlaffiliation{ind}{Independent Researcher}
\icmlcorrespondingauthor{Omer Haq}{haqomer1@gmail.com}

\icmlkeywords{language models, recurrent memory, latent state, sequential inference, uncertainty,
parameter-efficient adaptation}

\vskip 0.3in
]

\printAffiliationsAndNotice{}

\begin{abstract}
Pretrained language models condition on a finite token context, so evidence that leaves the context
can no longer influence prediction. We study Recursive State-Augmented Language Models (Recursive
SALM), in which a persistent latent state $z_t$ is updated after every query--response observation
$D_t=(x_t,y_t)$ and generates an adaptation of a frozen pretrained model, giving the state-conditioned
predictor $p_\theta(y_t\mid x_t,z_t)$. We derive the filtering recursion of this latent sequence model,
obtain deterministic recurrent state as its Dirac special case, and give a trajectory evidence lower
bound and a particle approximation for stochastic state. With frozen GPT-2 small and all earlier
evidence removed from the decoder context: (i)~a 64-dimensional state accumulates 4--16 evidence
statements over 16 hypotheses and trails a fine-tuned model that reads the full transcript by only
\V{v3/h3/diffabs}--\V{v3/h1/diffabs} accuracy points (five seeds), but falls \V{v3/h5/diffabs} points behind at
32 hypotheses and loses unrepeated clues beyond its training horizon; (ii)~among seven state-to-model
interfaces with near-equal parameter counts, a state-gated low-rank update (gated LoRA) is the most
accurate and the only one with no poorly trained seed; (iii)~on noisy evidence with an exact Bayesian
posterior, a four-particle stochastic SALM trained with that posterior as an auxiliary target reaches
\V{sto/sto/kl}~nats of KL divergence to it, against \V{sto/det/kl} for a matched deterministic SALM.
Both select posterior modes almost always, so the stochastic advantage lies in calibration, not
decisions. Calibrated inference without posterior supervision, larger hypothesis spaces and
natural-language memory remain open.
\end{abstract}

\section{Introduction}
\label{sec:intro}

An autoregressive language model predicts from the tokens in its context. In a long-running
interaction the history grows without bound, and keeping it visible makes attention and the
key--value cache grow with it. Truncation bounds the cost but discards evidence that a later
prediction may need. Two broad remedies exist. \emph{Context-based memory} keeps information in
token form and returns it to the context: longer windows, cached activations, or retrieval
\citep{dai2019transformerxl,rae2020compressive,wu2022memorizing,lewis2020rag}. \emph{Recurrent
memory} instead compresses the past into a learned representation that is carried forward, for
example as memory tokens or soft prompts
\citep{bulatov2022rmt,chevalier2023compress}. A third possibility, which we study here, is to let a
persistent latent state change the \emph{predictor itself}: after each observation, a learned
recurrence updates the state, and the state generates an adaptation of a frozen pretrained model
for the next prediction.

Recursive State-Augmented Language Models (Recursive SALM) formalize this idea. Each completed
observation $D_t=(x_t,y_t)$, a query $x_t$ and its response $y_t$, is read by an encoder that uses
the pretrained model's own representations, combined with the previous state by a learned
transition, and the resulting state $z_{t+1}$ is written into the model as an input-, activation-
or weight-level adaptation. Prediction uses the state-conditioned distribution
$p_\theta(y_t\mid x_t,z_t)$; the earlier observations need not appear in the decoder context at
all. The state may be deterministic, or it may carry a distribution over latent states that is
updated by Bayes' rule. SALM extends the latent-state predictor-adaptation direction of
LatentTrack~\citep{haq2026latenttrack}, which combined latent filtering with hypernetwork weight
generation for online prediction, to pretrained language models with interchangeable read and write
interfaces.

We organize the paper around three questions, each tied to a controlled experiment.

\textbf{Q1. Can a compact persistent state accumulate evidence across observations?}
Retaining a transcript and accumulating evidence are different operations. A transcript keeps
every observation and leaves the combination to the reader; a recurrent state must combine each
new observation with what it already holds, in a fixed number of dimensions, without access to the
raw history. We test this on \emph{hypothesis-code memory} (HCM): a hidden code is drawn from $M$
candidates, a sequence of truthful clues each eliminates some of them, and a final query asks for
the code (Section~\ref{sec:det_hcm}). Only the state connects the clues to the query.

\textbf{Q2. How should a latent state adapt a pretrained language model?}
The same state can be written into the model as a soft prompt, a prefix, a feature modulation,
cross-attention, or a low-rank weight update. We compare seven such interfaces at near-equal
trainable parameter counts and identify the one that is accurate and reliable across seeds
(Section~\ref{sec:adapters}).

\textbf{Q3. Can stochastic latent states represent uncertainty over competing hypotheses?}
When evidence is noisy, the correct output of a memory is a posterior over hypotheses, not a single
answer. A deterministic state is a point estimate of the filtering law and has no explicit
representation of uncertainty; a stochastic state can represent a distribution. We compare the two
on a noisy-evidence variant of HCM whose exact posterior is available for evaluation
(Section~\ref{sec:stochastic}), separating the accuracy of the selected hypothesis from the
calibration of the belief.

\paragraph{Contributions.}
\begin{enumerate}
\item A latent sequence formulation for adapting a pretrained predictor from sequential
query--response observations (Sections~\ref{sec:formulation}--\ref{sec:training}). It derives the
predictive and filtering recursions, shows that the deterministic recurrent state is the Dirac special
case and that evidence can then enter only through an observation-conditioned transition, and gives
a trajectory ELBO with a causal variational family and a particle approximation for stochastic state.
\item A read--transition--write architecture that separates observation encoding, latent dynamics,
and predictor adaptation, so that each can be varied while the others are held fixed
(Section~\ref{sec:architecture}).
\item A five-seed study of evidence accumulation in a 64-dimensional state on frozen GPT-2 small,
across evidence lengths and hypothesis-space sizes, with a full-transcript fine-tuned baseline,
state interventions and a test of retention beyond the training horizon
(Section~\ref{sec:det_hcm}).
\item A five-seed comparison of seven state-to-model interfaces that motivates gated LoRA, with a
ten-seed, 512-episode comparison against the transcript baseline (Section~\ref{sec:adapters}).
\item A five-seed comparison of stochastic and deterministic SALM against an exact Bayesian posterior,
which separates hypothesis selection from posterior calibration and states precisely where the exact
posterior is used in training (Section~\ref{sec:stochastic}).
\end{enumerate}

The evidence is synthetic and controlled by design: the tasks admit exact posteriors and make the
information path from evidence to prediction unambiguous. A preliminary natural-language
experiment on MultiRC, in which SALM trails a full-context baseline and depends only weakly on
passage-specific state, is reported in Appendix~\ref{app:multirc} and is not a contribution of this
paper.

\section{Related Work}
\label{sec:related}

\paragraph{Context-based memory.}
Transformer-XL reuses hidden activations of the previous segment, and Compressive Transformers
compress older activations into a secondary memory
\citep{dai2019transformerxl,rae2020compressive}. Memorizing Transformers retrieve cached key--value
pairs, and retrieval-augmented generation retrieves text
\citep{wu2022memorizing,lewis2020rag}. In all of these the predictor reads stored representations
of past tokens, and its memory grows with the stored history or the retrieval store. A Recursive SALM
state has a fixed size and is a function of the history, not a copy of it. Our full-context
baseline is the limiting case of context-based memory in which the entire transcript is visible.

\paragraph{Recurrent memory representations.}
The Recurrent Memory Transformer passes dedicated memory tokens from one segment to the next
\citep{bulatov2022rmt}, and AutoCompressors summarize preceding segments as soft prompts that are
prepended to later ones \citep{chevalier2023compress}. These carry a learned recurrent
representation, but the representation enters the model as additional input positions. SALM's
soft-prompt and prefix interfaces belong to this family. SALM separates the recurrence from the
interface that uses its state, so that the same state can instead modulate activations or gate
low-rank weight updates, and Section~\ref{sec:adapters} compares these choices directly. Gated
recurrent networks and structured or selective state-space models
\citep{cho2014gru,hochreiter1997lstm,gu2022s4,gu2023mamba,dao2024mamba2} use state transitions as
sequence-modeling layers inside the network; in SALM the state process wraps a frozen pretrained
predictor and supplies its adaptations.

\paragraph{Latent-variable sequence models and filtering.}
Classical state-space models and hidden Markov models distinguish prediction from posterior
filtering \citep{kalman1960new,rabiner1989tutorial}, and particle filters approximate both with
weighted samples \citep{gordon1993novel}. Deep sequential latent-variable models, including the
VRNN, deep Kalman filters and stochastic recurrent layers, train such models with amortized
variational bounds \citep{chung2015vrnn,krishnan2015deep,krishnan2017structured,fraccaro2016srnn},
and filtering-based objectives train particle approximations directly
\citep{maddison2017fivo,naesseth2018vsmc,le2018aesmc}. Latent world models such as PlaNet and
Dreamer learn transitions that can be rolled out without new observations
\citep{hafner2019planet,hafner2020dreamer,hafner2023dreamerv3}. Recursive SALM borrows the
filtering decomposition but differs in what the latent variable does: it conditions a pretrained
language model's \emph{parameters or activations}, rather than being decoded by a model trained from
scratch, and its transition is conditioned on the completed observation
(Section~\ref{sec:regimes} contrasts this with autonomous latent dynamics).

\paragraph{Conditional and generated adaptation.}
Fast-weight networks and hypernetworks generate parameters of one network from another
\citep{schmidhuber1992fast,ha2017hypernetworks}; Bayesian hypernetworks place a distribution over
the generated weights \citep{krueger2017bayesian}. Parameter-efficient interfaces such as soft
prompts, prefixes, residual adapters, feature-wise modulation and LoRA
\citep{lester2021power,li2021prefix,houlsby2019adapter,perez2018film,hu2022lora} define the space in
which SALM writes. HyperTuning generates prefixes or LoRA from examples \citep{phang2023hypertuning};
Text-to-LoRA generates adapters from task descriptions, Doc-to-LoRA internalizes documents, and SHINE
maps supplied context to LoRA in a single pass
\citep{charakorn2025texttolora,charakorn2026doctolora,liu2026shine}. These methods are \emph{one-shot}:
an adapter is generated from a given input and is not updated by later observations. SALM's
adaptation is \emph{sequential}: each observation updates a state that persists, and the adaptation
at time $t$ depends on the whole history through that state.

\paragraph{Sequential latent-conditioned adaptation.}
The closest prior work generates adaptations from a recurrent or latent state. LatentTrack combines
latent filtering with hypernetwork weight generation for online prediction
\citep{haq2026latenttrack}; Recursive SALM builds directly on this direction and is not the first
latent-state approach to Bayesian predictor adaptation. Code2LoRA generates LoRA adapters for a frozen
code language model and, in its evolving variant, updates a GRU state from successive repository
diffs \citep{hotsko2026code2lora}. Infinite-Parameter LLMs propose online weight generation
conditioned on a latent belief and evaluate a categorical belief over codes derived from data
\citep{hu2026infinite}. Neural Processes condition predictions on a set of observations through a
latent variable \citep{garnelo2018neural}. Relative to these, our contributions are a common
filtering formulation for deterministic and stochastic state over query--response observations, a
controlled comparison of write interfaces under a fixed state process, and an evaluation that
measures both state dependence and, where an exact posterior exists, the calibration of the
represented belief.

\paragraph{Test-time adaptation and continual learning.}
Test-time training layers and Titans update memory parameters at inference by gradient steps on a
self-supervised loss \citep{sun2025ttt,behrouz2025titans}, and test-time adaptation methods such as
Tent minimize an unsupervised objective on incoming data \citep{wang2021tent}. Continual-learning
methods regularize parameter updates against forgetting \citep{kirkpatrick2017overcoming}. SALM's
inference-time adaptation is gradient-free: all learning happens in training, and at inference the
adaptation changes only through the learned transition. This makes each update cheap and
deterministic in cost, but it also limits adaptation to the range of the trained transition; SALM
does not address lifelong learning of new skills.

\paragraph{Summary of distinctions.}
(1)~Context-based memory stores past tokens or their activations; SALM stores a fixed-size state.
(2)~Recurrent memory representations usually enter as extra input positions; SALM uses the state to
condition the predictor through an interchangeable interface, including weight updates.
(3)~Hypernetwork adapters are generated once from a given context; SALM's adaptation is recurrent.
(4)~Most recurrent memories are deterministic; SALM's formulation admits a stochastic state with
Bayesian filtering, and we test when that representation improves the calibration of beliefs.

\section{Latent Sequence Formulation}
\label{sec:formulation}

Let $D_{1:T}=(D_1,\ldots,D_T)$ be a sequence of observations, where $t$ indexes an \emph{update
event}, not a token. In dialogue, $D_t=(x_t,y_t)$ comprises a query $x_t$ and its response $y_t$.
Let $z_{1:T}$ be latent states with $z_t\in\mathcal Z$. The state $z_t$ is available when predicting
at event $t$; the completed observation $D_t$ then enters the transition to $z_{t+1}$. We write
$D_{<t}$ and $z_{<t}$ for histories, $D_{\le t}$ and $z_{\le t}$ when the current element is
included.

\subsection{Chain rule and model assumptions}

Ordering the variables as $z_1,D_1,\ldots,z_T,D_T$, the chain rule gives the unrestricted
factorization
\begin{align}
p(D_{1:T},z_{1:T})
&=\prod_{t=1}^{T}p(z_t\mid z_{<t},D_{<t})\nonumber\\[-2pt]
&\hspace{3.2em}\times p(D_t\mid D_{<t},z_{\le t}).
\label{eq:exact_factorization}
\end{align}
First-order Recursive SALM restricts both factors.

\begin{assumption}[First-order recursive state]
\label{ass:recursive}
For $t\ge 2$, the transition depends only on the previous state and the previous completed
observation:
\begin{equation}
p(z_t\mid z_{<t},D_{<t})=p_\phi(z_t\mid z_{t-1},D_{t-1}).
\label{eq:recursive_assumption}
\end{equation}
\end{assumption}

\begin{assumption}[Current-state decoding]
\label{ass:decoding}
Given the observed history and the current state, the current observation is independent of
earlier states:
\begin{equation}
p(D_t\mid D_{<t},z_{\le t})=p(D_t\mid D_{<t},z_t).
\label{eq:decoder_assumption}
\end{equation}
\end{assumption}

The decoder may still read observed history directly. Writing
$\ell_t(z_t)=p(D_t\mid D_{<t},z_t)$ for the likelihood of the realized observation, the assumptions
yield
\begin{equation}
p(D_{1:T},z_{1:T})=p_\phi(z_1)\prod_{t=1}^{T}\ell_t(z_t)\prod_{t=2}^{T}p_\phi(z_t\mid z_{t-1},D_{t-1}),
\label{eq:recursive_joint}
\end{equation}
where $p_\phi(z_1)$ is the initial state law, and $p(D_{1:T})=\int p(D_{1:T},z_{1:T})\,dz_{1:T}$
(sums for discrete states). Model parameters are suppressed in joint and marginal probabilities when
the relevant factors are explicit.

\subsection{Turn-level model with supplied queries}

In passive SALM the query $x_t$ is supplied exogenously; only the response is modeled. Let
$c_t=W(D_{<t})$ be the token context that a window rule $W$ exposes to the decoder, and define the
\emph{response likelihood}
\begin{equation}
\lambda_t(z_t)=p_\theta(y_t\mid c_t,x_t,z_t).
\label{eq:response_likelihood}
\end{equation}
When no earlier observation is visible, $c_t=\emptyset$ and this is the state-conditioned predictor
$p_\theta(y_t\mid x_t,z_t)$ used throughout our experiments. For a fixed query sequence $x_{1:T}$,
the conditional complete-data law is
\begin{equation}
\begin{aligned}
p(y_{1:T},z_{1:T}\mid x_{1:T})={}&p_\phi(z_1)\prod_{t=1}^{T}\lambda_t(z_t)\\
&\times\prod_{t=2}^{T}p_\phi(z_t\mid z_{t-1},D_{t-1}).
\end{aligned}
\label{eq:conditional_trajectory}
\end{equation}
Each factor uses only the queries supplied up to its event, and
$p(y_{1:T}\mid x_{1:T})$ integrates Equation~\eqref{eq:conditional_trajectory} over $z_{1:T}$.
For a response $y_t=(y_{t,1},\ldots,y_{t,m_t})$ the likelihood is autoregressive,
\begin{equation}
p_\theta(y_t\mid c_t,x_t,z_t)=\prod_{j=1}^{m_t}p_\theta(y_{t,j}\mid c_t,x_t,y_{t,<j},z_t),
\label{eq:chunk_likelihood}
\end{equation}
with the same $z_t$ conditioning every token of the response.

\subsection{Predictive and filtering recursions}
\label{sec:prediction}

Let $b_t(z_t)=p(z_t\mid D_{<t})$ denote the \emph{predictive} state law, with $b_1=p_\phi(z_1)$. In
the turn-level model the response predictive and the exact \emph{filtering} posterior are
\begin{align}
p(y_t\mid x_t,D_{<t})&=\int\lambda_t(z_t)\,b_t(z_t)\,dz_t,
\label{eq:response_predictive}\\
p(z_t\mid D_{\le t})&=\frac{\lambda_t(z_t)\,b_t(z_t)}{p(y_t\mid x_t,D_{<t})},
\label{eq:response_filtering}
\end{align}
defined when the predictive likelihood is positive. The next predictive law propagates the filtered
law through the transition:
\begin{equation}
b_{t+1}(z_{t+1})=\int p_\phi(z_{t+1}\mid z_t,D_t)\,p(z_t\mid D_{\le t})\,dz_t.
\label{eq:recursive_belief}
\end{equation}
The joint observation model has the same recursion with $\ell_t$ in place of $\lambda_t$.
A completed observation therefore affects the belief through \emph{two} channels:
(a)~its likelihood reweights belief about the state that generated it
(Equation~\eqref{eq:response_filtering}), and (b)~it conditions the kernel that propagates that belief
to the next state (Equation~\eqref{eq:recursive_belief}). Which channels are available distinguishes
the three regimes below. A derivation of
Equations~\eqref{eq:response_predictive}--\eqref{eq:recursive_belief} from the assumptions is given in
Appendix~\ref{app:derivations}.

\subsection{Deterministic, autonomous and stochastic regimes}
\label{sec:regimes}

\paragraph{Deterministic recurrent SALM.}
A deterministic transition places a Dirac measure at the output of a state update $f_\phi$,
\begin{equation}
p_\phi(dz_{t+1}\mid z_t,D_t)=\delta_{f_\phi(z_t,D_t)}(dz_{t+1}),
\label{eq:dirac_transition}
\end{equation}
and the initial law is a point mass at $\bar z_1$.

\begin{proposition}[Deterministic filtering]
\label{prop:deterministic}
Under Equation~\eqref{eq:dirac_transition} and $b_1=\delta_{\bar z_1}$, for every history with
positive likelihood the predictive law $b_t$ and the filtering law $p(z_t\mid D_{\le t})$ are both the
point mass $\delta_{\bar z_t}$, where
\begin{equation}
\bar z_{t+1}=f_\phi(\bar z_t,D_t),
\label{eq:deterministic_recursion}
\end{equation}
and $p(y_t\mid x_t,D_{<t})=p_\theta(y_t\mid c_t,x_t,\bar z_t)$.
\end{proposition}

\emph{Proof.} By induction: if $b_t=\delta_{\bar z_t}$ and $\lambda_t(\bar z_t)>0$, then
Equation~\eqref{eq:response_filtering} reweights a single atom and returns $\delta_{\bar z_t}$, and
Equation~\eqref{eq:recursive_belief} pushes it to $\delta_{f_\phi(\bar z_t,D_t)}$.
Equation~\eqref{eq:response_predictive} evaluated at a point mass gives the last claim. $\square$

Channel~(a) is therefore inert in deterministic SALM: the likelihood can never move a point mass. All
evidence must enter through the observation-conditioned update $f_\phi$, which is why the transition
of Assumption~\ref{ass:recursive} reads $D_t$. The state is a deterministic function of the realized
history, and the response distribution is the language model's autoregressive distribution at that
state, from which tokens can still be sampled.

\paragraph{Autonomous latent dynamics.}
A latent dynamical system in the style of classical state-space models or latent world models uses an
\emph{autonomous} kernel $p_\phi(z_{t+1}\mid z_t)$ that does not read the observation; observations
enter only through the likelihood. This is the special case of Assumption~\ref{ass:recursive} in which
the kernel ignores $D_t$. Such a model can be rolled forward without observations, which an
observation-conditioned kernel cannot do without specified or simulated $D_t$. Combining the two cases
shows a limitation of autonomous deterministic dynamics:

\begin{proposition}[No evidence through an autonomous Dirac kernel]
\label{prop:autonomous}
If the kernel is autonomous and deterministic, $p_\phi(dz_{t+1}\mid z_t)=\delta_{g_\phi(z_t)}$, and
$b_1=\delta_{\bar z_1}$, then $b_t=\delta_{g_\phi^{t-1}(\bar z_1)}$ for every history with positive
likelihood: the state, and hence the prediction $p_\theta(y_t\mid c_t,x_t,z_t)$ through the state,
is independent of the observations.
\end{proposition}

\emph{Proof.} Neither channel can act: channel~(a) cannot move a point mass
(Proposition~\ref{prop:deterministic}), and channel~(b) is absent because $g_\phi$ does not read $D_t$.
$\square$

An autonomous model must therefore carry a non-degenerate belief for evidence to influence the state,
and then it relies entirely on Bayesian reweighting. Recursive SALM conditions its kernel on $D_t$,
which makes deterministic memory possible.

\paragraph{Stochastic Recursive SALM.}
A stochastic transition represents several states for the same history. With an observation encoding
$e_t$ of $D_t$, a diagonal Gaussian kernel takes the form
\begin{equation}
\begin{aligned}
(\mu_{t+1},\log\sigma_{t+1})&=F_\phi(z_t,e_t),\\
p_\phi(z_{t+1}\mid z_t,D_t)&=\mathcal N\!\big(z_{t+1};\mu_{t+1},\operatorname{diag}(\sigma_{t+1}^2)\big),
\end{aligned}
\label{eq:gaussian_transition}
\end{equation}
with reparameterized samples $z_{t+1}=\mu_{t+1}+\sigma_{t+1}\odot\epsilon_{t+1}$,
$\epsilon_{t+1}\sim\mathcal N(0,I)$ \citep{kingma2014vae,rezende2014stochastic}. Mixtures can
parameterize the same kernel. Both channels now act: the likelihood reweights hypotheses about the
state, and the observation moves the states themselves. Different states induce different
adaptations of the same backbone, and the response predictive
Equation~\eqref{eq:response_predictive} averages the adapted models' \emph{probabilities}, not their
states.

\begin{remark}[What a deterministic state can and cannot represent]
\label{rem:det_uncertainty}
A deterministic state is a point estimate of the filtering law; it has no explicit representation of
uncertainty \emph{over states}. Its predictive $p_\theta(y_t\mid x_t,\bar z_t)$ can still be a
non-degenerate distribution over responses, and a trained predictor may place calibrated probabilities
on competing answers. Nothing in the formulation guarantees this: calibration of a deterministic
SALM is an empirical property of the trained map, whereas a stochastic SALM represents uncertainty
structurally through its belief over states. Section~\ref{sec:stochastic} measures both against an exact
posterior.
\end{remark}

\subsection{Observation granularity}

The update index $t$ is separate from autoregressive token time. A chunking rule $\mathcal C$ may
partition a token stream $s_{1:N}$ as $(D_1,\ldots,D_T)=\mathcal C(s_{1:N})$; token, fixed-block and
semantic chunking expose different amounts of evidence to each transition. In our experiments one
completed query--response pair is one observation. The final query of each episode is
\emph{read-only}: it is answered from the current state and does not update it.

\section{Read--Transition--Write Architecture}
\label{sec:architecture}

\begin{figure}[t]
\centering
\begin{tikzpicture}[
  font=\scriptsize, >={Stealth[length=4pt]},
  box/.style={draw, rounded corners=2pt, minimum height=7mm, minimum width=14mm, inner sep=2pt, align=center},
  st/.style={draw, circle, minimum size=8mm, inner sep=0pt, fill=blue!8},
  llm/.style={box, fill=gray!12},
  io/.style={inner sep=1pt}]
\node[st]  (z)  at (0,0)    {$z_t$};
\node[box] (G)  at (1.75,0) {write $G_\eta$};
\node[llm] (M)  at (3.85,0) {frozen LM\\[-1pt]$\theta_0$ adapted by $A_t$};
\node[io]  (x)  at (3.85,1.0) {query $x_t$};
\node[io]  (y)  at (5.75,0) {$y_t$};
\node[st]  (z2) at (0,-1.55)    {$z_{t+1}$};
\node[box] (F)  at (1.75,-1.55) {transition $F_\phi$};
\node[llm] (E)  at (3.85,-1.55) {read $E_\psi$\\[-1pt](frozen LM)};
\node[io]  (D)  at (5.75,-1.55) {$D_t{=}(x_t,y_t)$};
\draw[->] (z) -- (G);
\draw[->] (G) -- node[above]{$A_t$} (M);
\draw[->] (x) -- (M);
\draw[->] (M) -- (y);
\draw[->] (y) -- (D);
\draw[->] (D) -- (E);
\draw[->] (E) -- node[above]{$e_t$} (F);
\draw[->] (z) -- (F);
\draw[->] (F) -- (z2);
\end{tikzpicture}
\caption{One Recursive SALM update. The write module turns the state $z_t$ into an adaptation $A_t$
of the frozen language model, which predicts $y_t$ from the query $x_t$ alone. After the response, the
read module encodes the completed observation $D_t=(x_t,y_t)$ with the frozen model, and the transition
combines the encoding with $z_t$. Earlier observations never enter the decoder context.}
\label{fig:architecture}
\end{figure}
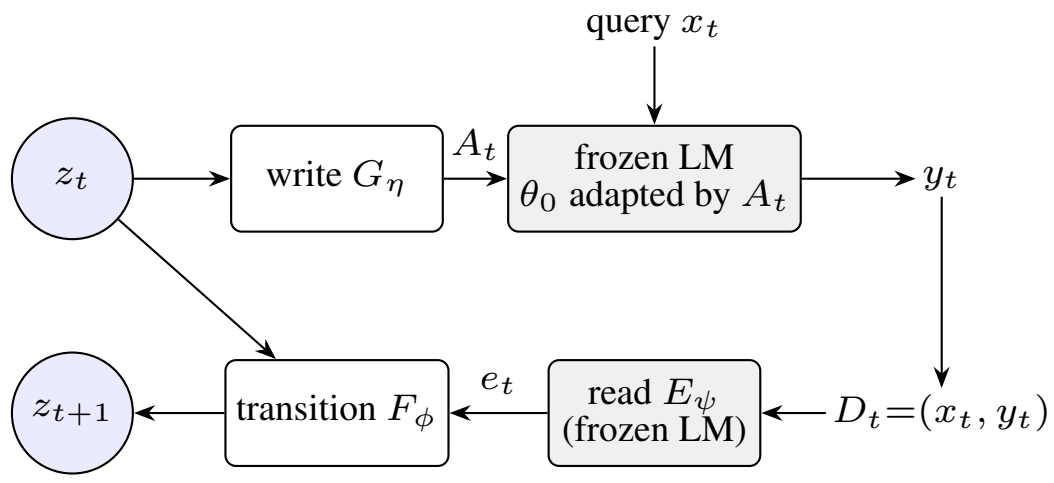

At event $t$, the write module maps the available state to an adaptation $A_t$; the decoder uses
$A_t$ with the visible context $c_t$ and query $x_t$ to predict $y_t$. Once $D_t$ is complete, the read
module encodes it and the transition propagates the state (Figure~\ref{fig:architecture}). For
deterministic SALM,
\begin{equation}
\begin{aligned}
e_t&=E_\psi\big(\{H^{(\ell)}_{\theta_0}(D_t)\}_{\ell\in\mathcal S}\big),\\
\bar z_{t+1}&=F_\phi(\bar z_t,e_t),\qquad A_{t+1}=G_\eta(\bar z_{t+1}).
\end{aligned}
\label{eq:read_transition_write}
\end{equation}
Here $\theta_0$ are the frozen backbone weights, $H^{(\ell)}_{\theta_0}(D_t)$ the layer-$\ell$
representations of the tokenized observation and $\mathcal S$ the selected layers. The trainable
parameters $\psi$, $\phi$ and $\eta$ belong to the read encoder, transition and write module. We write
$p_\theta$ for the state-conditioned predictor, comprising the frozen backbone and the write module;
the map $f_\phi$ of Equation~\eqref{eq:deterministic_recursion} is the composition of the read encoder
with the cell $F_\phi$. In the stochastic model, $F_\phi$ outputs the parameters of
Equation~\eqref{eq:gaussian_transition}.

\paragraph{Read.} The encoder converts token representations of the observation into a vector, by
mean, last-token or learned attention pooling. In our experiments it reads the final layer of the
frozen, \emph{unadapted} backbone on the current observation alone.

\paragraph{Transition.} A GRU, recurrent MLP or state-space update can implement $F_\phi$; its state
dimension may differ from the encoding dimension. We use a GRU cell \citep{cho2014gru}.

\paragraph{Write.} The write module $G_\eta$ has parameters shared across events; only its input, the
state, changes. We consider three levels at which the adaptation can act:
\begin{itemize}
\item \emph{Input and attention prefixes.} A \textbf{soft prompt} prepends generated embeddings to the
input \citep{lester2021power}; \textbf{prefix tuning} supplies generated keys and values to every
attention layer \citep{li2021prefix}; \textbf{deep prompting} inserts generated prompt tokens at every
layer.
\item \emph{Activations.} \textbf{FiLM} applies $h'=\gamma(z_t)\odot h+\beta(z_t)$ to hidden
representations \citep{perez2018film}; a \textbf{cross-attention} sidecar lets the decoder attend to a
projection of the state.
\item \emph{Weights.} A low-rank update changes a base operator $W_0$ to $W_0+\Delta W_t$
\citep{hu2022lora}. \textbf{Generated LoRA} emits both factors from the state with a hypernetwork
\citep{ha2017hypernetworks}. \textbf{Gated LoRA} keeps learned, shared factors and lets the state gate
their rank components:
\begin{equation}
\Delta W_s(z_t)=\frac{\alpha}{r}\,A_s\,\operatorname{diag}\!\big(\tanh(G_s z_t+c_s)\big)\,B_s,
\label{eq:gated_lora}
\end{equation}
for each adapted site $s$, with $A_s\in\mathbb R^{d_{\mathrm{in}}\times r}$,
$B_s\in\mathbb R^{r\times d_{\mathrm{out}}}$ and a linear gate map $G_s\in\mathbb R^{r\times d_z}$.
The state selects a point in an $r$-dimensional family of updates spanned by learned directions,
rather than synthesizing the directions themselves.
\end{itemize}
These interfaces differ in the size of the generated object, where in the network it acts, and how
directly it changes the computation on the current query.

\subsection{Finite context and persistent state}
\label{sec:context}

The response predictor of Equation~\eqref{eq:response_likelihood} has two paths from history: the
token context $c_t=W(D_{<t})$ and the state. An observation that $W$ omits can still affect a later
prediction through the state; how well it survives depends on the state size, the transition, the
training objective and interference from later observations.

\paragraph{Storage and computation.} For a dense Transformer over $n$ tokens there are $\Theta(n^2)$
query--key interactions per layer; with key--value caching, decoding one token costs $\Theta(n)$
interactions per layer, and the cache grows linearly in $n$ \citep{kwon2023pagedattention}. A
persistent state $z_t\in\mathbb R^d$ stores $\mathcal O(d)$ numbers between updates, independent of
how many observations preceded it; a $K$-particle approximation stores $\mathcal O(Kd)$ coordinates
plus weights. This bound concerns storage only. Each update requires encoding the observation and
applying the transition, and each prediction requires generating and applying the adaptation, so at
short horizons SALM can be slower than reading a short transcript (Appendix~\ref{app:compute}).

\section{Training and Inference}
\label{sec:training}

\subsection{Deterministic SALM}

By Proposition~\ref{prop:deterministic}, the deterministic likelihood is the product of
$\lambda_t(\bar z_t)$ with $\bar z_t$ computed from $D_{<t}$. Maximum likelihood on responses supervised
at events $\mathcal T_{\mathrm{sup}}$ minimizes
\begin{equation}
\mathcal L_{\mathrm{det}}=-\sum_{t\in\mathcal T_{\mathrm{sup}}}\log p_\theta(y_t\mid c_t,x_t,\bar z_t),
\label{eq:deterministic_loss}
\end{equation}
which by Equation~\eqref{eq:chunk_likelihood} is token-level cross-entropy on the supervised
responses. Unsupervised observations still affect the loss through the state trajectory. Truncated
backpropagation through time would detach gradients at a window boundary,
$z_t\leftarrow\operatorname{stopgrad}(z_t)$, while keeping the numerical state; the decoder context
determines which tokens are visible, and the gradient window determines how far training credit
propagates through the state. In all our experiments each episode is a single gradient window.

\subsection{Stochastic SALM: trajectory ELBO}

Consider the turn-level model with queries $x_{1:T}$ and responses with $p(y_{1:T}\mid x_{1:T})>0$.
For a variational law $q_\rho(z_{1:T}\mid D_{1:T})$ absolutely continuous with respect to the exact
trajectory posterior and with finite expectations, the evidence lower bound is
\begin{equation}
\begin{aligned}
\mathcal J_{\mathrm{ELBO}}=\E_{q_\rho}\Big[&\log p(y_{1:T},z_{1:T}\mid x_{1:T})\\
&-\log q_\rho(z_{1:T}\mid D_{1:T})\Big],
\end{aligned}
\label{eq:trajectory_elbo}
\end{equation}
and $\log p(y_{1:T}\mid x_{1:T})-\mathcal J_{\mathrm{ELBO}}=\KL\big(q_\rho\,\Vert\,p(z_{1:T}\mid D_{1:T})\big)$,
the divergence to the smoothing posterior. A causal, ancestor-conditioned family is
\begin{equation}
q_\rho(z_{1:T}\mid D_{1:T})=q_\rho(z_1\mid D_1)\prod_{t=2}^{T}q_\rho(z_t\mid z_{t-1},D_t).
\label{eq:causal_q}
\end{equation}
Its $t$th factor can read the completed $D_t$ to infer $z_t$, whereas the generative transition into
$z_t$ reads $D_{t-1}$. With local divergences
$K_1=\KL(q_\rho(z_1\mid D_1)\Vert p_\phi(z_1))$ and, for $t\ge2$,
\begin{equation}
K_t(z_{t-1})=\KL\big(q_\rho(z_t\mid z_{t-1},D_t)\,\Vert\,p_\phi(z_t\mid z_{t-1},D_{t-1})\big),
\label{eq:local_kl}
\end{equation}
substituting Equation~\eqref{eq:conditional_trajectory} gives
\begin{equation}
\begin{aligned}
\mathcal J_{\mathrm{ELBO}}={}&\sum_{t=1}^{T}\E_{q_\rho}[\log\lambda_t(z_t)]\\
&-K_1-\sum_{t=2}^{T}\E_{q_\rho(z_{t-1})}[K_t(z_{t-1})],
\end{aligned}
\label{eq:recursive_elbo_expanded}
\end{equation}
where $q_\rho(z_{t-1})$ is the corresponding marginal of Equation~\eqref{eq:causal_q}
(Appendix~\ref{app:derivations}). The local terms compare conditional laws for the same ancestor. The
trajectory ELBO is distinct from an online objective that substitutes an approximate predictive
marginal at each step.

\subsection{Particle approximation}
\label{sec:particles}

Equations~\eqref{eq:response_predictive}--\eqref{eq:recursive_belief} can be approximated with $K$
weighted particles $\{z_t^{(k)},w_t^{(k)}\}$. A \emph{bootstrap} filter
\citep{gordon1993novel} uses the transition as its proposal. At event $t$ it predicts with the mixture
\begin{equation}
p(y_t\mid x_t,D_{<t})\approx\sum_{k=1}^{K}w_{t-1}^{(k)}\,\lambda_t\big(z_t^{(k)}\big),
\label{eq:mc_prediction}
\end{equation}
then applies Bayes' rule to the weights,
$w_t^{(k)}\propto w_{t-1}^{(k)}\lambda_t(z_t^{(k)})$, and propagates each particle,
$z_{t+1}^{(k)}\sim p_\phi(\cdot\mid z_t^{(k)},D_t)$. Without resampling, the weight update is exactly
channel~(a) and the propagation is channel~(b). The logarithm of the product of per-event mixture
likelihoods is a lower bound on $\log p(y_{1:T}\mid x_{1:T})$ in expectation, which filtering variational
objectives maximize \citep{maddison2017fivo,naesseth2018vsmc,le2018aesmc}.

\subsection{Diagnosing state use}
\label{sec:diagnostics}

A pretrained decoder can predict well from visible tokens while ignoring the state, so performance
alone does not establish memory. Holding the query and visible context fixed, we replace the state
used for prediction: the \emph{zero} (initial-state) intervention removes episode-specific information
while keeping the write module active, and the \emph{shuffled} intervention substitutes another
episode's state through a derangement, which preserves the marginal distribution of states while
breaking their association with the history. For evidence accumulation we additionally remove the
first observation of an episode, and we \emph{reset} or \emph{swap} the whole state after the first
observation and let the remaining observations update it. Performance changes under these
interventions measure dependence on the state; a probe that decodes an earlier attribute from $z_t$
measures what the state contains.

\section{Experimental Protocol}
\label{sec:exp_setup}

All experiments test one information path: earlier observations can influence the final prediction
\emph{only} through the persistent state. This section describes what is shared; each experiment then
states its task before its results.

\paragraph{Model.} The backbone is GPT-2 small (124M parameters, Hugging Face revision \texttt{607a30d})
\citep{radford2019gpt2}, frozen throughout; no backbone parameter is in any optimizer. The read encoder
applies learned attention pooling to the final-layer token states of the frozen, unadapted GPT-2 on the
current observation alone, followed by \emph{centered RMS normalization}: the mean pooled vector over
the training observations (a fixed buffer computed once per run from training data) is subtracted
before RMS normalization. Without centering, pooled GPT-2 representations of different evidence
sentences were nearly collinear in pilot runs and the state barely moved. The transition is a GRU cell
with a 64-dimensional state and $\bar z_1=0$ (16-dimensional in Section~\ref{sec:stochastic}). SALM's
decoder receives the current query and \emph{no earlier observation}, $c_t=\emptyset$, so the
predictor is exactly $p_\theta(y_t\mid x_t,z_t)$.

\paragraph{Full-context baseline.} A static LoRA (rank 64, $\alpha=128$) on the same twelve fused
attention projections of the same frozen GPT-2 (2{,}359{,}296 trainable parameters) is fine-tuned on
the complete preceding transcript followed by the query. It has the same supervision and checkpoint
selection as SALM, and it routes evidence through the token context instead of a state. Its
interventions remove the transcript or replace it with another episode's.

\paragraph{Supervision.} Evidence observations update the state and carry no loss. Only the answer
token of the final, read-only query is supervised, so training minimizes
Equation~\eqref{eq:deterministic_loss} restricted to that token, with backpropagation through the
entire episode.

\paragraph{Metrics.} For a query with candidate set $\mathcal H$ we score
$s(h)=\log p_\theta(y=h\mid x,z)$ and normalize over candidates,
$q_\theta(h)=\exp s(h)/\sum_{h'\in\mathcal H}\exp s(h')$. \emph{Accuracy} is the fraction of queries
with $\arg\max_h q_\theta(h)=h^\star$. \emph{Oracle KL} is $\KL(\pstar\Vert q_\theta)$ to the exact
posterior $\pstar$ given the episode's evidence; it equals the candidate NLL $-\log q_\theta(h^\star)$
when $\pstar$ is one-hot. Section~\ref{sec:stochastic} adds calibration metrics.

\paragraph{Statistics.} A \emph{seed} is an independent training initialization; within an experiment
all seeds and arms use identical data and test episodes, verified by dataset fingerprints. We report
means $\pm$ sample standard deviations over seeds. Differences between arms are paired by seed, and
their 95\% intervals are Student-$t$ intervals over seeds, which reflect seed-level variation. Every
model is evaluated once on its test split at the checkpoint selected on validation data. Every number
in Sections~\ref{sec:det_hcm}--\ref{sec:stochastic} was recomputed from saved per-episode predictions
and checked against the values recorded by the original training runs.

\section{Q1: Evidence Accumulation in a Deterministic State}
\label{sec:det_hcm}

\subsection{Task: hypothesis-code memory}

An episode draws a hidden codeword uniformly from a fixed pool of $M$ numeric codes. It then presents
$T$ evidence statements, each a hard, truthful predicate on the codeword from three families: parity
(``The codeword is even.''), threshold comparison, and inclusive range (``The codeword is between 21
and 34.''). Each predicate has one canonical wording. A final read-only query asks for the codeword,
and the answer token is the only supervised token. Episodes are accepted only if the full evidence
identifies exactly one code, so the exact posterior $\pstar$ is one-hot at the query, while earlier
prefixes of the evidence leave several codes consistent. Training, validation and test splits contain
512, 128 and 512 episodes with balanced targets, and their evidence \emph{sets} are disjoint: no
unordered set of clues is shared between splits. Individual clue sentences do recur across splits, so
the task tests the combination of known evidence statements into unseen evidence sets, not
generalization to new wording.

We vary the number of evidence statements and the size of the hypothesis space in five cells:
H1 $(T,M)=(4,16)$, H2 $(8,16)$, H3 $(16,16)$, H4 $(8,8)$ and H5 $(8,32)$. In SALM, the $T$ statements
pass through the 64-dimensional state one at a time and never enter the decoder context; at the query
the decoder sees only the query and the adaptation generated from $z_{T+1}$.
Full-context LoRA reads all $T$ statements and the query in its context.

\paragraph{Training.} SALM uses gated LoRA (Equation~\eqref{eq:gated_lora}; rank 43, $\alpha=86$,
2{,}370{,}180 trainable parameters, matched to the baseline's 2{,}359{,}296; the interface choice is
examined in Section~\ref{sec:adapters}). Both arms use AdamW (weight decay 0.01, gradient clipping 1,
batch 8), float32 throughout, validation every epoch, at least 10 and at most 30 epochs with patience 8,
and select the checkpoint with maximum validation accuracy. SALM uses learning rate $10^{-3}$ and the
baseline $10^{-4}$, the setting each arm had been developed with; neither was tuned on these cells. We
report five seeds per arm and cell (50 runs on Kaggle T4 GPUs); seeds 3 and 4 were trained later from a
byte-identical frozen copy of the original source and configuration.

\subsection{Results}

\begin{table*}[t]
\caption{Evidence accumulation (Q1). Test accuracy on 512 episodes per cell, mean $\pm$ s.d.\ over five
seeds; chance is $1/M$. $\Delta$: SALM minus full-context LoRA, paired by seed, with a 95\% $t$-interval
over seeds. \emph{First clue removed}: accuracy when the first evidence statement is deleted from the
episode (SALM: from the state's input stream; full context: from the transcript). \emph{Shuffle
penalty}: increase in SALM's answer NLL when its final state is replaced by another test episode's
state (mean over seeds).}
\label{tab:det_hcm}
\centering\small
\begin{tabular}{lrrcccccc}
\toprule
 & & & \multicolumn{2}{c}{Accuracy (\%)} & & \multicolumn{2}{c}{First clue removed (\%)} & Shuffle \\
\cmidrule(lr){4-5}\cmidrule(lr){7-8}
Cell & $T$ & $M$ & SALM & Full context & $\Delta$ [95\% CI] & SALM & Full context & penalty (nats) \\
\midrule
H1 & 4 & 16 & 92.0$\pm$3.8 & 96.4$\pm$0.7 & $-$4.4 [$-$9.1, +0.4] & 43.4$\pm$1.8 & 33.7$\pm$0.8 & 24.4 \\
H2 & 8 & 16 & 89.6$\pm$3.4 & 93.8$\pm$0.3 & $-$4.3 [$-$8.3, $-$0.2] & 42.7$\pm$1.1 & 35.4$\pm$2.4 & 23.5 \\
H3 & 16 & 16 & 91.2$\pm$3.7 & 95.2$\pm$1.5 & $-$4.0 [$-$8.5, +0.5] & 45.4$\pm$2.5 & 29.8$\pm$0.8 & 23.2 \\
H4 & 8 & 8 & 97.2$\pm$1.2 & 96.4$\pm$0.7 & +0.7 [$-$0.2, +1.6] & 36.4$\pm$1.4 & 46.4$\pm$1.7 & 21.9 \\
H5 & 8 & 32 & 73.4$\pm$2.4 & 94.2$\pm$0.5 & $-$20.9 [$-$24.1, $-$17.6] & 31.8$\pm$5.8 & 24.7$\pm$2.9 & 20.4 \\
\bottomrule
\end{tabular}

\end{table*}

\begin{figure*}[t]
\centering
\includegraphics[width=\textwidth]{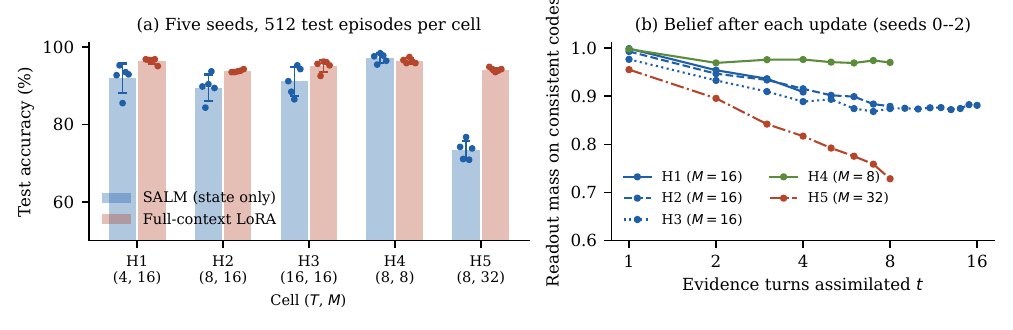}
\caption{(a) Test accuracy per cell; bars are seed means, dots individual seeds, error bars $\pm$ s.d.
(b) Evidence accumulation inside the state: after each update, the recall readout applied to the
current state, and the probability mass it places on codes still consistent with the evidence seen so
far (mean over seeds 0--2, \V{ret/episodes} test episodes per seed).}
\label{fig:det_hcm}
\end{figure*}

\paragraph{A compact state accumulates evidence over 4--16 updates.}
With $M=16$ hypotheses, SALM reaches \V{v3/h1/salm}\%, \V{v3/h2/salm}\% and \V{v3/h3/salm}\% at
$T=4$, 8 and 16, against \V{v3/h1/full}\%, \V{v3/h2/full}\% and \V{v3/h3/full}\% for full-context LoRA
(Table~\ref{tab:det_hcm}, Figure~\ref{fig:det_hcm}a). The paired gap is \V{v3/h1/diff}, \V{v3/h2/diff}
and \V{v3/h3/diff} points, with intervals \V{v3/h1/diffci}, \V{v3/h2/diffci} and \V{v3/h3/diffci}. It
does not grow with the number of updates over the trained horizons: a state updated sixteen times is as
accurate as one updated four times. The baseline keeps every statement verbatim and combines them only
when the query arrives; SALM must fold each statement into the same 64 numbers as it arrives, without
access to earlier statements, and it recovers most of the transcript model's accuracy in doing so.

\paragraph{The hypothesis space, not the horizon, limits accuracy.}
At $M=8$ SALM matches the baseline (\V{v3/h4/salm}\% against \V{v3/h4/full}\%, $\Delta=\V{v3/h4/diff}$
\V{v3/h4/diffci}). At $M=32$ it trails by \V{v3/h5/diffabs} points (\V{v3/h5/salm}\% against
\V{v3/h5/full}\%, $\Delta$ interval \V{v3/h5/diffci}), while the baseline barely changes. This gap is
not under-training: one H5 run continued under a sealed 60-epoch copy of its configuration stopped by
patience at epoch \V{v3/h5x/last} without improving on its epoch-\V{v3/h5x/best} checkpoint. With the
same state size and the same number of updates (H2, H4, H5 all have $T=8$), accuracy falls as the number
of hypotheses the state must discriminate grows.

\paragraph{Predictions depend on the state, and early evidence is carried in it.}
Replacing SALM's final state with another episode's raises its answer NLL by
\V{v3/shufpen/min}--\V{v3/shufpen/max} nats in every cell. Deleting the first evidence statement lowers
accuracy to \V{v3/early/salmmin}--\V{v3/early/salmmax}\% for SALM and to
\V{v3/early/fullmin}--\V{v3/early/fullmax}\% for the baseline, so both depend on the earliest evidence.
For SALM that evidence can reach the query only through $T-1$ subsequent state updates. Interventions
inside the recursion confirm where it is kept (seeds 0--2): resetting the state to $\bar z_1$ after the
first statement and letting the remaining statements update it lowers terminal accuracy to
\V{ret/reset/min}--\V{ret/reset/max}\%, and swapping in another episode's state after the first
statement lowers it to \V{ret/swap/min}--\V{ret/swap/max}\%. After the swap the readout still places
\V{ret/late/min}--\V{ret/late/max} of its mass on codes consistent with the episode's own later
statements: the model combines the foreign early evidence with the new evidence rather than
overwriting it. A cross-validated linear probe on the state decodes the bits of the first statement
that later statements leave undetermined with balanced accuracy of at least \V{ret/d1min} at every
update on H1--H4, falling to \V{ret/h5/d1min} on H5.

\paragraph{Errors accumulate during filtering.}
Applying the recall readout after each update (Figure~\ref{fig:det_hcm}b) shows how the belief evolves.
On H4 the readout places \V{ret/h4/cm1} of its mass on still-consistent codes after the first statement
and \V{ret/h4/cmT} after the last; on H5 this mass falls gradually from \V{ret/h5/cm1} to
\V{ret/h5/cmT}. Readout mass leaks onto eliminated codes a little at each update, and the terminal
error is the accumulation of these leaks rather than a failure of the final readout.

\begin{figure}[t]
\centering
\includegraphics[width=\columnwidth]{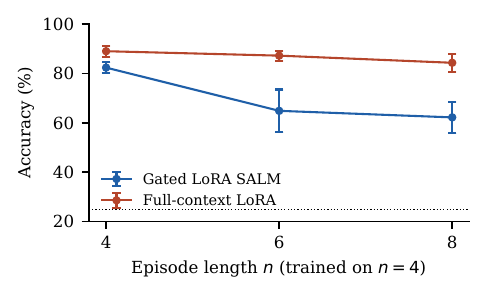}
\caption{Retention beyond the training horizon. Four-turn-trained models on longer episodes in which the
first clue is necessary (condition~B); mean $\pm$ s.d.\ over ten seeds, \V{lh/n} episodes per point.
Dotted: uniform guess.}
\label{fig:long_horizon}
\end{figure}

\paragraph{Retention beyond the training horizon fails.}
The cells above train and test at the same $T$. To test retention beyond it, we evaluate four-turn
models (three clues and a query, trained on the four-code task of Section~\ref{sec:adapters}) on longer
episodes, without retraining. In condition~B the first clue is \emph{necessary}: without it at least two
codes remain, so the answer is recoverable only if the state keeps that clue through all later updates.
Over ten seeds and \V{lh/n} shared episodes per length, gated LoRA SALM falls from \V{lh/B4/gated}\% at
four turns to \V{lh/B6/gated}\% at six and \V{lh/B8/gated}\% at eight, while full-context LoRA stays at
\V{lh/B4/full}\%, \V{lh/B6/full}\% and \V{lh/B8/full}\% (Figure~\ref{fig:long_horizon}; gap at eight
turns \V{lh/B8/diff} points, \V{lh/B8/diffci}). When the added clues are redundant (condition~A, seeds
0--2), SALM stays within six points of the baseline up to twelve turns (Appendix~\ref{app:long_horizon}). Combined with
the stable H1--H3 results, this locates the limitation: a state trained on a given horizon accumulates
evidence reliably within it, but an unrepeated clue is partly overwritten by updates beyond it.

\section{Q2: How Should the State Adapt the Model?}
\label{sec:adapters}

\subsection{Setup}

\paragraph{Task.} The comparison uses the four-code version of HCM on which the write interfaces were
originally developed. A hidden code is drawn uniformly from $\{44,19,93,90\}$ (single GPT-2 tokens).
Three evidence turns follow, each a hard, truthful statement of a parity, threshold or inclusive-range
predicate, possibly negated (``My codeword is greater than 58.''), and a read-only query asks for the
code. An episode is kept only if the three clues identify exactly one code, at least two codes remain
after the first clue, and the last clue alone leaves at least two. The 256 training, 64 validation and
64 test episodes share no evidence sequence, although individual clue sentences recur. The test targets
are imbalanced (\V{ad/testcounts} of \V{ad/n}), so a constant prediction scores up to
\V{ad/majority}\%; predicting the second most frequent code scores \V{ad/zeroconst}\%.

\paragraph{Interfaces.} All SALM arms share the read encoder and the GRU transition
(Section~\ref{sec:exp_setup}) and differ only in the write module (Section~\ref{sec:architecture}).
\emph{Generated LoRA}: a hypernetwork emits rank-8 ($\alpha=16$) factors for the twelve fused attention
projections; the slice producing $B_t$ is zero-initialized so the untrained model is exactly GPT-2.
\emph{Gated LoRA}: Equation~\eqref{eq:gated_lora} on the same twelve sites, rank 43, $\alpha=86$.
\emph{Soft prompt}: a one-hidden-layer network maps $z_t$ to four prompt embeddings. \emph{FiLM},
\emph{prefix tuning} (eight key/value prefix tokens per layer) and \emph{deep prompting} (eight prompt
tokens per layer) have one capacity setting each, chosen so that the total trainable parameter count
approximates generated LoRA's 2{,}355{,}520; all lie within 0.7\% of it
(Table~\ref{tab:params} in Appendix~\ref{app:adapter_details}). \emph{Cross-attention} is outside the
match: its 768-wide attention module alone exceeds the budget, giving 41\% more trainable parameters.

\paragraph{Training.} All arms, including full-context LoRA, use AdamW (learning rate $10^{-4}$, weight
decay 0.01, clipping 1, batch 8), validate every two epochs, stop by patience on validation answer NLL
after at least 12 epochs, and select the minimum-validation-NLL checkpoint. Runs that ended at the epoch
cap rather than by patience were continued under a rule fixed before any test evaluation (cap raised to
72 epochs, extended once to 100, same stopping rule); this applied to some FiLM, prefix, deep-prompt and
cross-attention runs (marked $^\ast$ in Table~\ref{tab:adapter}). Every arm has seeds 0--4; gated LoRA
and full-context LoRA also have seeds 5--9. Training ran on CPU.

\subsection{Results}

\begin{table*}[t]
\caption{Write-interface comparison on four-code HCM (Q2): test accuracy (\%) on \V{ad/n} episodes.
\emph{Train h}: mean CPU training hours over seeds 0--2 (seeds 3--4 ran under a different thread
configuration and are excluded from timing). \emph{Acc.}: mean $\pm$ s.d.\ over five seeds;
\emph{Worst}: lowest seed. \emph{Zero}/\emph{Shuffled}: the state at the query replaced by the initial
state or by another episode's state (for full context: transcript removed or shuffled). $\Delta$ vs.\
FC: paired difference to full-context LoRA, 95\% $t$-interval over five seeds. $^\ast$Some seeds
continued under the pre-registered epoch-cap rule.}
\label{tab:adapter}
\centering\small
\setlength{\tabcolsep}{4pt}
\begin{tabular}{lrrcccccc}
\toprule
Write interface & Params & Train h & Acc.\ (5) & Per seed (0--4) & Worst & Zero & Shuffled & $\Delta$ vs.\ FC [95\% CI] \\
\midrule
Generated LoRA & 2.36M & 0.71 & 74.7$\pm$15.0 & 86, 80, 72, 86, 50 & 50 & 25.6$\pm$8.9 & 30.6$\pm$5.0 & $-$13.1 [$-$31.8, +5.6] \\
Gated LoRA & 2.37M & 0.91 & 88.4$\pm$3.2 & 91, 84, 92, 86, 89 & 84 & 31.2$\pm$0.0 & 34.1$\pm$2.0 & +0.6 [$-$2.9, +4.1] \\
Soft prompt & 2.35M & 0.90 & 82.5$\pm$9.1 & 92, 81, 89, 81, 69 & 69 & 31.2$\pm$0.0 & 36.2$\pm$1.7 & $-$5.3 [$-$16.3, +5.6] \\
FiLM$^\ast$ & 2.36M & 1.56 & 46.6$\pm$19.3 & 44, 31, 80, 44, 34 & 31 & 31.2$\pm$0.0 & 31.9$\pm$4.6 & $-$41.2 [$-$64.4, $-$18.1] \\
Prefix tuning$^\ast$ & 2.35M & 2.61 & 55.6$\pm$16.1 & 80, 53, 55, 56, 34 & 34 & 31.2$\pm$0.0 & 27.8$\pm$7.2 & $-$32.2 [$-$52.2, $-$12.1] \\
Deep prompt$^\ast$ & 2.36M & 2.10 & 83.8$\pm$6.1 & 88, 88, 88, 83, 73 & 73 & 31.2$\pm$0.0 & 30.9$\pm$2.0 & $-$4.1 [$-$11.4, +3.3] \\
Cross-attention$^\ast$ & 3.32M & 1.53 & 78.1$\pm$24.7 & 94, 88, 84, 91, 34 & 34 & 17.5$\pm$9.4 & 35.0$\pm$2.6 & $-$9.7 [$-$40.3, +20.9] \\
\midrule
Full-context LoRA & 2.36M & 0.32 & 87.8$\pm$0.7 & 88, 88, 89, 88, 88 & 88 & 25.6$\pm$8.9 & 34.7$\pm$1.3 & -- \\
\bottomrule
\end{tabular}

\end{table*}

\begin{figure}[t]
\centering
\includegraphics[width=\columnwidth]{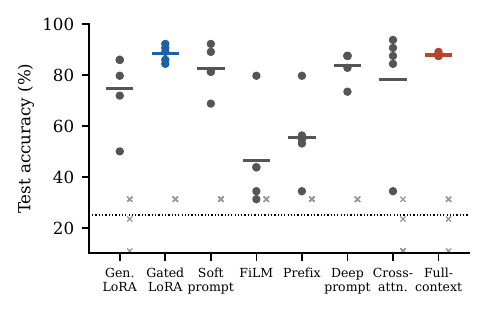}
\caption{Per-seed test accuracy of each interface on four-code HCM (dots; bar: mean over five seeds).
Grey crosses: the same models with the state zeroed. Dotted: uniform guess.}
\label{fig:adapter}
\end{figure}

\paragraph{Gated LoRA is the most accurate and the most reliable interface.}
Gated LoRA reaches \V{ad/gatedlora/acc}\% over five seeds, and its worst seed reaches
\V{ad/gatedlora/worst}\% (Table~\ref{tab:adapter}, Figure~\ref{fig:adapter}). Every other interface has
at least one seed that trains poorly: soft prompt (\V{ad/softprompt/acc}\%, worst \V{ad/softprompt/worst}\%),
deep prompting (\V{ad/deepprompt/acc}\%, worst \V{ad/deepprompt/worst}\%), generated LoRA
(\V{ad/generatedlora/acc}\%, worst \V{ad/generatedlora/worst}\%) and cross-attention
(\V{ad/crossattention/acc}\%, worst \V{ad/crossattention/worst}\%, despite 41\% more parameters). FiLM
(\V{ad/film/acc}\%) and prefix tuning (\V{ad/prefix/acc}\%) remain far below, even after continued
training. With three seeds, gated LoRA, soft prompt, deep prompting and cross-attention had been
indistinguishable (\V{ad/gatedlora/acc3}, \V{ad/softprompt/acc3}, \V{ad/deepprompt/acc3} and
\V{ad/crossattention/acc3}\%); seeds 3 and 4 separate them by reliability rather than by typical
accuracy. On this 64-episode split gated LoRA is statistically indistinguishable from full-context LoRA
($\Delta=\V{ad/gatedlora/diff}$ \V{ad/gatedlora/diffci}). Its training cost is similar to soft
prompting and below the prefix-style interfaces, though about three times that of the transcript
baseline on CPU (\V{ad/gatedlora/hours} against \V{ad/fullcontextlora/hours}~h per run).

\paragraph{Every interface uses the state.}
Replacing the state by its initial value reduces every SALM interface to between
\V{ad/crossattention/zero}\% and \V{ad/gatedlora/zero}\% on average, at or below the constant-prediction
level, and substituting another episode's state reduces it to \V{ad/prefix/shuf}--\V{ad/softprompt/shuf}\%.
For cross-attention the zero-state accuracy falls below the constant level because the sidecar remains
active and adapts the model away from its unadapted behaviour.

\paragraph{Ten seeds and a larger test set.}
The 64-episode split cannot resolve differences of a few points (one binomial standard error is about
four points). We therefore evaluated the ten gated-LoRA and ten full-context runs, each once at its
selected checkpoint, on an independent set of \V{lt/n} four-turn episodes from the same generator, none
of which occurs in the original splits. Gated LoRA reaches \V{lt/gated/acc}\% and full-context LoRA
\V{lt/full/acc}\%, a paired difference of \V{lt/diff/acc} points (95\% $t$-interval over seeds
\V{lt/diff/accci}; a hierarchical bootstrap over seeds and episodes gives \V{lt/diff/acchb}). Oracle KL
is \V{lt/gated/kl} against \V{lt/full/kl} nats. The transcript model is therefore more accurate and
better calibrated on this task, and SALM stays within about \V{lt/diff/accabs} points while carrying all
evidence in its 64-dimensional state. Zeroing or shuffling the state reduces gated LoRA to
\V{lt/gated/zero}\% and \V{lt/gated/shuf}\% (no seed above \V{lt/gated/maxshuf}\%), close to the 25\% of uniform guessing.

\paragraph{Why gated LoRA?}
We did not run experiments that isolate the cause, but the comparison is consistent with a simple
account. Gated LoRA learns its low-rank directions as ordinary shared parameters and lets the state
choose only a bounded gate vector, so the state controls the predictor through a smooth,
low-dimensional and well-conditioned map. Generated LoRA must synthesize the directions themselves from
the state, and input-level interfaces must compete with the query tokens for attention. We use gated
LoRA for all other experiments; the comparison covers one backbone, one task family and one parameter
budget and does not establish that gated LoRA is preferable in general.

\section{Q3: Stochastic State and Uncertainty over Hypotheses}
\label{sec:stochastic}

The tasks above have one-hot posteriors, so a correct memory needs only to identify one hypothesis. We
now make the evidence unreliable, so that the correct output of the memory is a posterior distribution,
and ask whether a stochastic state represents that posterior better than a deterministic one.

\subsection{Task and exact posterior}

A hidden code $C$ is drawn uniformly from the four codes $\{44,19,93,90\}$. Each of three observations
$D_t=(x_t,y_t)$ pairs a question $x_t$ about a fixed predicate of the code (parity, or a threshold
between adjacent codes), chosen independently of $C$, with a yes/no answer $y_t$ that reports the
predicate's true value with probability $\rho=0.8$, independently across observations given $C$. The
reliability is never stated in the text. A final read-only query asks for the code. Writing
$\pi_t(c)=\rho$ if the answer $y_t$ is correct for code $c$ and $1-\rho$ otherwise, the exact posterior
after $t$ observations is
\begin{equation}
\pstar(C=c\mid D_{1:t})=\frac{\prod_{s=1}^{t}\pi_s(c)}{\sum_{c'}\prod_{s=1}^{t}\pi_s(c')},
\label{eq:noisy_posterior}
\end{equation}
which we compute in closed form for every episode. Its mean entropy after three observations is
\V{sto/postent}~nats on the test split (uniform: $\log 4=1.39$), so most episodes stay ambiguous, and in
\V{sto/tied}\% of test episodes two or more codes share the maximal posterior probability. Training
and validation use 512 and 128 episodes from which 20 unordered answer patterns are held out; we test on
a natural split of \V{sto/n} episodes and on an \emph{unseen-pattern} split of \V{sto/nu} episodes that
contains only the held-out patterns.

\subsection{Models}

\paragraph{Stochastic SALM.} The state is 16-dimensional. The transition is the Gaussian kernel of
Equation~\eqref{eq:gaussian_transition} with a residual mean, $\mu_{t+1}=0.9\,z_t+0.1\,\mathrm{GRU}(e_t,z_t)$,
and a learned scale (initial value 0.1). The initial law is a fixed four-component Gaussian mixture, and
the $K=4$ particles are drawn one per component in a random order. Inference is the bootstrap particle
filter of Section~\ref{sec:particles} without resampling: each particle is propagated by the transition
and its weight is multiplied by the response likelihood $\lambda_t(z_t^{(k)})=p_\theta(y_t\mid x_t,z_t^{(k)})$
of the observed answer. Each particle generates its own gated-LoRA adaptation of GPT-2. At the query,
particle $k$ defines a candidate distribution $q_k(c)=p_\theta(C=c\mid x_{\mathrm{query}},z^{(k)})$, normalized
over the four codes, and the prediction is the mixture $q_{\mathrm{mix}}=\sum_kw^{(k)}q_k$, which mixes
probabilities rather than states.

\paragraph{Deterministic SALM.} The matched deterministic model has the same encoder, transition cell
and write module, a Dirac transition, a point initial state and $K=1$ (\V{sto/det/params} against
\V{sto/sto/params} trainable parameters).

\paragraph{Objective, and where the exact posterior is used.} Both models minimize
\begin{equation}
\begin{aligned}
\mathcal L={}&\underbrace{-\log q_{\mathrm{mix}}(C)}_{\mathcal L_{\mathrm{recall}}}
\;\underbrace{-\frac1T\sum_{t=1}^{T}\log\hat p(y_t\mid x_t,D_{<t})}_{\mathcal L_{\mathrm{evidence}}}\\
&\underbrace{-\sum_c\pstar(c\mid D_{1:T})\log q_{\mathrm{mix}}(c)}_{\mathcal L_{\mathrm{oracle}}},
\end{aligned}
\label{eq:stochastic_objective}
\end{equation}
where $\hat p$ is the particle estimate of Equation~\eqref{eq:mc_prediction} (with $K=1$ the
deterministic likelihood). $\mathcal L_{\mathrm{recall}}$ uses the sampled code, an observed label;
$\mathcal L_{\mathrm{evidence}}$ is a filtering-variational term on the answers
\citep{maddison2017fivo}. $\mathcal L_{\mathrm{oracle}}$ is a cross-entropy to the \emph{exact posterior}
of Equation~\eqref{eq:noisy_posterior}. The exact posterior is therefore used as a training target, and
checkpoints are selected by validation KL to it; it never enters the state or the decoder. \textbf{Both
models are oracle-supervised}: the comparison asks which state representation can \emph{express} a
posterior when trained toward it, not whether either can \emph{infer} one from answers alone. A variant
that adds an optimal-transport term between particle weights and the posterior (Appendix~\ref{app:sto_details})
gives the same result. Training uses AdamW (learning rate $3\times10^{-4}$, batch 4) and stops after six
epochs without a 0.002-nat validation improvement (10 to 60 epochs). We train five seeds of each model;
each is evaluated once on the test splits.

\paragraph{Metrics.} Besides accuracy against the sampled code and oracle KL, we report the NLL of the
sampled code, the Brier score \citep{brier1950verification}, the expected calibration error of the top
prediction against the sampled code over ten bins \citep{guo2017calibration}, and a \emph{posterior
calibration error}, $\E\,\big|\max_cq(c)-\pstar(\hat c)\big|$ with $\hat c=\arg\max_cq(c)$, which compares the
model's confidence with the exact probability that its chosen code is correct. The exact posterior itself
provides the Bayes-optimal reference for every metric.

\subsection{Results}

\begin{table*}[t]
\caption{Stochastic against deterministic SALM on noisy four-code HCM (Q3), natural test split
(\V{sto/n} episodes) unless stated; mean $\pm$ s.d.\ over five seeds. Both models are trained with the exact
posterior as an auxiliary target. Sto.\ $-$ det.: paired difference, 95\% $t$-interval over seeds.
\emph{Exact posterior}: Bayes-optimal reference. Accuracy is measured against the sampled code; with
$\rho=0.8$ and three reports even the exact posterior is correct in only \V{sto/bayes/acc}\% of episodes.
Mixing gain: increase in expected NLL under $\pstar$ when the particle mixture is replaced by a single
adaptation at the weighted mean state.}
\label{tab:stochastic}
\centering\small
\setlength{\tabcolsep}{4pt}
\begin{tabular}{lccccc}
\toprule
Metric (natural test split) & Stochastic & Deterministic & Sto.\ $-$ det.\ [95\% CI] & Exact posterior & Uniform \\
\midrule
Oracle KL (nats) $\downarrow$ & 0.009$\pm$0.002 & 0.042$\pm$0.007 & $-$0.032 [$-$0.039, $-$0.025] & 0.000 & 0.411 \\
NLL of true code $\downarrow$ & 0.980$\pm$0.002 & 1.011$\pm$0.009 & $-$0.030 [$-$0.041, $-$0.020] & 0.970 & 1.386 \\
Brier score $\downarrow$ & 0.540$\pm$0.001 & 0.555$\pm$0.005 & $-$0.015 [$-$0.020, $-$0.010] & 0.534 & 0.750 \\
ECE (10 bins) $\downarrow$ & 0.063$\pm$0.018 & 0.048$\pm$0.012 & +0.015 [$-$0.015, +0.045] & 0.037 & -- \\
Posterior calibration error $\downarrow$ & 0.035$\pm$0.006 & 0.081$\pm$0.010 & $-$0.045 [$-$0.058, $-$0.033] & 0.000 & -- \\
Accuracy (\%) $\uparrow$ & 53.9$\pm$0.8 & 55.4$\pm$0.6 & $-$1.6 [$-$2.9, $-$0.3] & 55.5 & 25.0 \\
Agreement with posterior mode (\%) & 100.0$\pm$0.0 & 99.8$\pm$0.1 & -- & 100.0 & -- \\
\midrule
Oracle KL, unseen patterns & 0.009$\pm$0.002 & 0.033$\pm$0.007 & $-$0.024 [$-$0.033, $-$0.015] & 0.000 & 0.556 \\
Mixing gain (nats) & 1.89$\pm$0.29 & 0 (by construction) & -- & -- & -- \\
\bottomrule
\end{tabular}

\end{table*}

\begin{figure*}[t]
\centering
\includegraphics[width=0.85\textwidth]{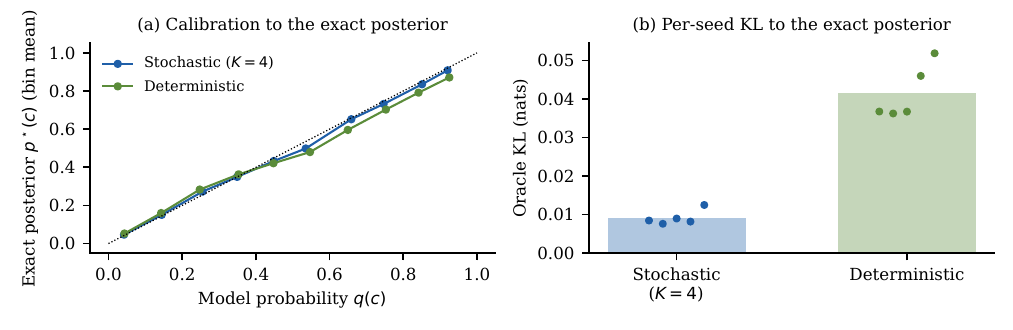}
\caption{(a) Calibration to the exact posterior: candidate probabilities of all test episodes and seeds,
binned by model probability, against the mean exact posterior probability of the same candidates
(bins with fewer than 20 entries omitted). (b) Oracle KL per seed.}
\label{fig:stochastic}
\end{figure*}

\paragraph{The stochastic state represents the posterior more faithfully.}
The stochastic model reaches oracle KL \V{sto/sto/kl}~nats against \V{sto/det/kl} for the deterministic
model (paired difference \V{sto/diff/kl}, \V{sto/diff/klci}), removing \V{sto/sto/fracremoved}\% and
\V{sto/det/fracremoved}\% of a uniform prediction's divergence (\V{sto/unif/kl}~nats). Every seed of the
stochastic model scores below every seed of the deterministic one (Figure~\ref{fig:stochastic}b). The
advantage holds on held-out answer patterns (\V{sto/sto/klu} against \V{sto/det/klu}; difference
\V{sto/diff/klu}, \V{sto/diff/kluci}). It is also visible in proper scoring rules against the sampled code:
NLL \V{sto/sto/nll} against \V{sto/det/nll} (exact posterior \V{sto/bayes/nll}) and Brier score
\V{sto/sto/brier} against \V{sto/det/brier} (\V{sto/bayes/brier}). The posterior calibration error is
\V{sto/sto/post_cal_err} against \V{sto/det/post_cal_err} (difference \V{sto/diff/post_cal_err},
\V{sto/diff/post_cal_errci}). Binned against the exact posterior (Figure~\ref{fig:stochastic}a), the
deterministic model's probabilities are too extreme at both ends: high probabilities exceed the exact
posterior by up to \V{cal/det/over} and low probabilities fall short of it by up to \V{cal/det/under},
whereas the stochastic model's largest deviation in any bin is \V{cal/sto/maxabs}, and most of its bins
lie close to the diagonal. The deterministic model is nonetheless far
from uninformed: with a single state it already removes most of the divergence, consistent with
Remark~\ref{rem:det_uncertainty}.

\paragraph{Selecting the right hypothesis is not the same as calibrated inference.}
Accuracy does not separate the models in a meaningful way. Both choose a code of maximal exact posterior
probability in nearly every episode (\V{sto/sto/mapagree}\% stochastic, \V{sto/det/mapagree}\%
deterministic), so their accuracy against the sampled code differs almost entirely through how ties among
equally probable codes are broken, which affects \V{sto/tied}\% of test episodes. The stochastic model is
\V{sto/diff/accabs} points less accurate on the natural split (\V{sto/diff/accci}) and \V{sto/diff/accuabs}
points more accurate on unseen patterns (\V{sto/diff/accuci}); neither difference reflects the quality of
the belief. The label-based expected calibration error is likewise inconclusive
(\V{sto/sto/ece} against \V{sto/det/ece}, difference \V{sto/diff/ececi}): with 512 noisy labels it is
dominated by sampling noise, which is why we also measure calibration against the exact posterior. A model
can thus be Bayes-accurate in its decisions while its probabilities are measurably miscalibrated.

\paragraph{How the particles represent the posterior.}
The mixture is essential to the stochastic model's prediction. Replacing it with a single adaptation at
the weighted mean state raises the expected NLL by \V{sto/sto/gain}~nats, and evaluating the same trained
models with a single particle gives oracle KL \V{sto/sto/klK1}, far worse than uniform, whereas eight
particles give \V{sto/sto/klK8} (Appendix~\ref{app:sto_details}). Each particle has specialized to one
code: the four particles predict four different codes (\V{sto/sto/distinctmin} distinct top codes per
test episode on average, in every seed), each with a mean top probability of at least
\V{sto/sto/pmaxmin}. Over an episode the particle states move only \V{sto/sto/disp} of their initial separation
(\V{sto/initdist}), whereas the learned per-report weight updates correlate at
\V{sto/sto/cohmin}--\V{sto/sto/cohmax} with the updates that each particle's own code prediction implies
under the true reliability. Flipping every answer changes the predicted distribution by an $\ell_1$
distance of \V{sto/sto/flippredmin}--\V{sto/sto/flippredmax}, against \V{sto/flip/oracle} for the exact
posterior, while moving the particle states by only \V{sto/sto/flipstatemin}--\V{sto/sto/flipstatemax}.
The stochastic model thus implements Bayesian filtering by \emph{reweighting a learned bank of four
hypotheses}, using channel~(a) of Section~\ref{sec:prediction}. The deterministic model must use
channel~(b): the same flip moves its state by \V{sto/det/flipstatemin}--\V{sto/det/flipstatemax} and
changes its prediction by \V{sto/det/flippredmin}--\V{sto/det/flippredmax}.

\paragraph{Cost and scope.}
The stochastic model needs \V{sto/inferratio}$\times$ the deterministic model's inference time per episode
(\V{sto/sto/infer}~s against \V{sto/det/infer}~s on CPU) and \V{sto/trainratio}$\times$ its training time
(seeds 0--2). The task is favourable to the particle design: four particles initialized one per component
face exactly four hypotheses, so reweighting fixed particles suffices, and the exact posterior supervises
the readout. These results show that a particle-based SALM \emph{can express} a near-exact posterior
through a frozen language model and that this improves calibration over a single state; they do not
show that stochastic SALM \emph{infers} calibrated posteriors without posterior supervision, or that the
advantage persists when hypotheses outnumber particles. Both questions are open
(Section~\ref{sec:discussion}).

\paragraph{Preliminary oracle-free results.}
Appendix~\ref{app:oracle_free} reports an earlier, separately run study in which the stochastic model is
trained \emph{without} the exact posterior: no $\mathcal L_{\mathrm{oracle}}$ term, and checkpoints selected on
validation NLL of the sampled code. That study uses natural-language report statements as evidence and has
three seeds per arm, so its numbers are not comparable with Table~\ref{tab:stochastic}.
\begin{itemize}
\item \textbf{Fit without the oracle.} With $\mathcal L_{\mathrm{recall}}$ alone, or with
$\mathcal L_{\mathrm{recall}}+\mathcal L_{\mathrm{evidence}}$, four particles reach oracle KL \V{of/A/kl} and
\V{of/B/kl} nats on its four-code task (uniform: \V{of/unif4/kl}).
\item \textbf{Particle diversity.} The particles still predict \V{of/A/distinct_particle_maps} and
\V{of/B/distinct_particle_maps} distinct codes per episode on average, out of four. Distinct particle
hypotheses that the weights select among therefore emerge without posterior supervision.
\item \textbf{The oracle term still matters.} Adding $\mathcal L_{\mathrm{oracle}}$ on the same protocol lowers
oracle KL to \V{of/C/kl} (difference \V{of/diff/C-B/kl}, \V{of/diff/C-B/klci}), so supervision still accounts
for much of the fit.
\item \textbf{Scaling.} With eight codes and four particles, the oracle-free model removes only
\V{of/M8K4/fracremoved}\% of the uniform divergence. Doubling to eight particles on four codes leaves oracle
KL unchanged within noise (\V{of/diff/M4K8-B/kl}, \V{of/diff/M4K8-B/klci}).
\end{itemize}
These preliminary results show that the particle structure does not depend on the oracle target. They do
not show calibrated posterior inference without it.

\section{Discussion}
\label{sec:discussion}

\paragraph{Accumulating evidence versus retaining a transcript.}
The full-context baseline answers by reading every clue at query time; its memory is the transcript, and
combining the evidence is deferred to the reader. SALM must combine each clue with its current state as
the clue arrives, using a learned update and a fixed number of dimensions. The deterministic experiments
show that this incremental combination works over 4--16 updates for up to 16 hypotheses with a
64-dimensional state, at a cost of about four accuracy points relative to the transcript, and that the
predictions depend on the state rather than on the query alone. They also show its two costs: accuracy
falls as the number of hypotheses grows at fixed state size, and a clue that later evidence does not
repeat is partly lost when the state is updated more often than in training. The per-update readout shows
the mechanism of both: a small amount of probability leaks onto eliminated hypotheses at each update.

\paragraph{The write interface matters.}
Under one state process and near-equal parameter budgets, interfaces that look similar on three seeds
differ sharply in reliability. Gated LoRA, which lets the state select among learned low-rank directions,
was the only interface without a poorly trained seed. Because the write module determines how a state is
read out, comparisons of recurrent memories that fix a single interface conflate the quality of the memory
with the quality of its readout.

\paragraph{Deterministic states and Bayesian beliefs.}
A deterministic state is a point estimate of the filtering law (Proposition~\ref{prop:deterministic}), and
its predictive distribution is calibrated only to the extent that training makes it so
(Remark~\ref{rem:det_uncertainty}). On the noisy task, a deterministic SALM trained toward the exact
posterior came close to it, but a particle-based state was measurably closer and better calibrated, while
both selected Bayes-optimal codes. The two learned different mechanisms: the deterministic state moves with
the evidence (channel~(b)), whereas the particles stay nearly fixed and are reweighted by the answer
likelihoods (channel~(a)). Evaluating such models by accuracy alone would have hidden both the calibration
difference and the mechanism.

\paragraph{Gradient-free inference-time adaptation.}
At inference SALM adapts a frozen model through forward computation only: no gradients, no optimizer
state, and a constant-size state between observations (\V{cmp/statekb}~KB in the deterministic
experiments). At the short horizons studied this does not make inference cheaper; SALM answers
\V{cmp/ratiomin}--\V{cmp/ratiomax}$\times$ more slowly than the transcript model on CPU
(Appendix~\ref{app:compute}), because every update re-encodes an observation and applies the adaptation.
Whether the constant-size state yields a practical advantage at long horizons was not measured.

\paragraph{Natural language (preliminary).}
In a four-seed MultiRC reading-comprehension experiment (Appendix~\ref{app:multirc}), SALM with gated
LoRA reaches held-out NLL \V{mrc/salm/nll} against \V{mrc/fc/nll} for full-context LoRA. Its predictions
depend strongly on having a state (zero-state NLL \V{mrc/salm/zeronll}) but hardly on having the
\emph{correct passage's} state (shuffled-state change \V{mrc/shuf}~nats, interval \V{mrc/shufci}). The
cause of this weak passage-specific dependence has not been established, and we do not interpret MultiRC
as evidence of natural-language memory.

\section{Limitations}
\label{sec:limitations}

\begin{itemize}
\item \textbf{Synthetic tasks, one backbone.} All central results use synthetic hypothesis-inference tasks
and frozen GPT-2 small. The deterministic tasks test the combination of evidence sentences that also occur
in training into unseen evidence sets. In an earlier design with paraphrased evidence, in which most test
sentences were unseen in training, deterministic SALM fitted its training set but generalized poorly;
generalization across wording is not established.
\item \textbf{Horizon and capacity.} Training horizons reach 16 updates, and models do not retain an
unrepeated clue beyond their training horizon. Accuracy falls at 32 hypotheses with a 64-dimensional state.
Capacity as a function of state size, overwrite and interference, and learning under truncated
backpropagation were not studied.
\item \textbf{Baseline protocol.} The full-context baseline and SALM share data, supervision and selection,
but in the deterministic HCM cells they use different learning rates that were not jointly tuned.
\item \textbf{Stochastic SALM.} The stochastic comparison uses four hypotheses, four particles initialized
one per hypothesis, one reliability, and an auxiliary loss on the exact posterior. It does not establish
oracle-free posterior inference or behaviour when hypotheses outnumber particles. The oracle-free and
eight-hypothesis results of Appendix~\ref{app:oracle_free} are preliminary (three seeds, a different evidence
format) and do not settle either question. Particles barely move, so the experiment tests reweighting
of a learned hypothesis bank rather than evidence-driven particle dynamics.
\item \textbf{Ranking and generation.} The first version of this preprint reported a two-turn task with
codewords unseen in training, in which several interfaces ranked the correct codeword above other candidates
well above chance but rarely generated it from the full vocabulary. We do not revisit that task; reading out
novel content verbatim from the state remains an open problem.
\item \textbf{Statistics.} Seed-level intervals with five seeds are wide, and the 64-episode adapter test
split resolves only large differences; the ten-seed, 512-episode evaluation covers only gated LoRA and the
baseline.
\end{itemize}

\section{Conclusion}
\label{sec:conclusion}

Recursive SALM couples observation-conditioned latent dynamics to the adaptation of a pretrained language
model. Its filtering recursion separates the two ways an observation can change a belief, by reweighting
hypotheses about the state and by moving the state, and shows that a deterministic recurrent state relies
entirely on the second. In controlled experiments with frozen GPT-2, a 64-dimensional deterministic state
accumulated up to sixteen evidence statements over sixteen hypotheses to within about four points of a
model that reads the whole transcript, and its predictions depended on that state; a state-gated low-rank
update was the most reliable way to write the state into the model; and when evidence was noisy, a
particle-based stochastic state represented the exact posterior more faithfully than a deterministic one,
improving calibration without changing decisions. These results establish persistent latent-conditioned
adaptation as a working mechanism for sequential inference beyond the decoder context, and they identify
what it has not yet shown: retention beyond the training horizon, scaling to larger hypothesis spaces,
calibrated inference without posterior supervision, and memory for natural language. It is a step toward
models that adapt continually from their own interaction history, not a solution to lifelong learning.

\bibliography{references}
\bibliographystyle{icml2024}

\clearpage
\appendix
\onecolumn
\section{Derivations}
\label{app:derivations}

Throughout, densities are with respect to a common dominating measure on $\mathcal Z$ (sums for discrete
states), and all conditional laws are evaluated at the realized observations.

\subsection{Predictive and filtering recursions}

\paragraph{Prediction.} Marginalizing $z_t$ and applying Assumption~\ref{ass:decoding},
\[
p(y_t\mid x_t,D_{<t})=\int p(y_t\mid x_t,D_{<t},z_t)\,p(z_t\mid x_t,D_{<t})\,dz_t
=\int\lambda_t(z_t)\,b_t(z_t)\,dz_t .
\]
The second equality uses $p(z_t\mid x_t,D_{<t})=p(z_t\mid D_{<t})=b_t(z_t)$: in the turn-level model the
query is supplied exogenously, and the complete-data law of Equation~\eqref{eq:conditional_trajectory}
contains no factor in which $z_t$ depends on $x_t$.

\paragraph{Filtering.} By Bayes' rule with the same two facts,
\[
p(z_t\mid D_{\le t})=\frac{p(y_t\mid x_t,D_{<t},z_t)\,p(z_t\mid D_{<t})}{p(y_t\mid x_t,D_{<t})}
=\frac{\lambda_t(z_t)\,b_t(z_t)}{p(y_t\mid x_t,D_{<t})}.
\]

\paragraph{Propagation.} Marginalizing $z_t$ and applying Assumption~\ref{ass:recursive},
\[
b_{t+1}(z_{t+1})=\int p(z_{t+1}\mid z_t,D_{\le t})\,p(z_t\mid D_{\le t})\,dz_t
=\int p_\phi(z_{t+1}\mid z_t,D_t)\,p(z_t\mid D_{\le t})\,dz_t .
\]
The first factor reduces to the transition because, in Equation~\eqref{eq:conditional_trajectory}, $z_{t+1}$
depends on the past only through $(z_t,D_t)$. Equations~\eqref{eq:response_predictive}--\eqref{eq:recursive_belief}
follow. For the joint observation model replace $\lambda_t$ by $\ell_t$.

\subsection{Trajectory ELBO and its gap}

For any $q_\rho$ with $q_\rho\ll p(z_{1:T}\mid D_{1:T})$,
\[
\log p(y_{1:T}\mid x_{1:T})
=\E_{q_\rho}\!\left[\log\frac{p(y_{1:T},z_{1:T}\mid x_{1:T})}{q_\rho(z_{1:T}\mid D_{1:T})}\right]
+\E_{q_\rho}\!\left[\log\frac{q_\rho(z_{1:T}\mid D_{1:T})}{p(z_{1:T}\mid D_{1:T})}\right],
\]
because $p(y_{1:T},z_{1:T}\mid x_{1:T})=p(y_{1:T}\mid x_{1:T})\,p(z_{1:T}\mid D_{1:T})$. The first term is
$\mathcal J_{\mathrm{ELBO}}$ and the second is $\KL(q_\rho\Vert p(z_{1:T}\mid D_{1:T}))\ge 0$.

\subsection{Expanded ELBO for the causal family}

Substituting Equations~\eqref{eq:conditional_trajectory} and~\eqref{eq:causal_q},
\[
\log\frac{p(y_{1:T},z_{1:T}\mid x_{1:T})}{q_\rho(z_{1:T}\mid D_{1:T})}
=\sum_{t=1}^{T}\log\lambda_t(z_t)
+\log\frac{p_\phi(z_1)}{q_\rho(z_1\mid D_1)}
+\sum_{t=2}^{T}\log\frac{p_\phi(z_t\mid z_{t-1},D_{t-1})}{q_\rho(z_t\mid z_{t-1},D_t)} .
\]
Take expectations under $q_\rho$. The second term gives $-K_1$. For $t\ge2$, the tower property over
$z_{t-1}$, with $z_t\sim q_\rho(\cdot\mid z_{t-1},D_t)$ conditionally on $z_{t-1}$, gives
\[
\E_{q_\rho}\!\left[\log\frac{p_\phi(z_t\mid z_{t-1},D_{t-1})}{q_\rho(z_t\mid z_{t-1},D_t)}\right]
=-\E_{q_\rho(z_{t-1})}\big[K_t(z_{t-1})\big],
\]
which is Equation~\eqref{eq:recursive_elbo_expanded}. The local divergence compares the variational and
generative conditionals \emph{for the same ancestor} $z_{t-1}$, while the variational factor reads $D_t$ and
the generative factor reads $D_{t-1}$.

\subsection{Particle estimate of the response likelihood}

With the bootstrap proposal and no resampling, each trajectory $z_{1:T}^{(k)}$ is a draw from the
generative transitions and the unnormalized weight after $T$ events is $\prod_t\lambda_t(z_t^{(k)})$. Hence
\[
\hat p(y_{1:T}\mid x_{1:T})=\frac1K\sum_{k=1}^{K}\prod_{t=1}^{T}\lambda_t\big(z_t^{(k)}\big)
=\prod_{t=1}^{T}\sum_{k=1}^{K}w_{t-1}^{(k)}\lambda_t\big(z_t^{(k)}\big)
\]
with normalized weights $w_{t-1}^{(k)}$ ($w_0^{(k)}=1/K$), and
$\E[\hat p(y_{1:T}\mid x_{1:T})]=p(y_{1:T}\mid x_{1:T})$ when the initial particles are i.i.d.\ from
$p_\phi(z_1)$. Under a stratified initial draw from an equal-weight mixture, as in
Section~\ref{sec:stochastic}, each particle is still marginally distributed as $p_\phi(z_1)$ and the
estimate remains unbiased by linearity. By Jensen's inequality,
$\E[\log\hat p]\le\log p(y_{1:T}\mid x_{1:T})$, so $\mathcal L_{\mathrm{evidence}}$ in
Equation~\eqref{eq:stochastic_objective} is (up to the factor $1/T$) the negative of a stochastic lower bound
on the response log-likelihood \citep{maddison2017fivo}. The per-event factors are the mixture predictions
of Equation~\eqref{eq:mc_prediction}.

\section{Deterministic Experiments: Details and Additional Results}
\label{app:det_details}

\subsection{Per-seed results for evidence accumulation}

Table~\ref{tab:det_hcm_seeds} lists every run behind Table~\ref{tab:det_hcm}. All runs of a cell share the
same data, verified by split fingerprints, and every model was evaluated once on the test split at the
checkpoint with maximum validation accuracy. Several SALM runs used all 30 epochs, and two (H2 seeds 0
and 3) selected their final epoch, so some SALM runs may have been stopped before convergence; the
continuation of one H5 run (Section~\ref{sec:det_hcm}) found no further improvement, but the other cells
were not continued. Seeds 0--2 of H1 and seed 0 of H2--H5 ran on the original source archive; all other
runs used a revision that adds cells and recovery paths but leaves the training path and the cell
configurations unchanged (identical configuration hashes).

\begin{table}[h]
\caption{Evidence accumulation: test accuracy (\%) of each seed on 512 episodes, and selected epochs.}
\label{tab:det_hcm_seeds}
\centering\small
\begin{tabular}{llccccccc}
\toprule
Cell & Arm & Seed 0 & Seed 1 & Seed 2 & Seed 3 & Seed 4 & Mean $\pm$ s.d. & Selected epochs \\
\midrule
H1 & SALM & 92.8 & 95.3 & 85.5 & 93.6 & 93.0 & 92.0$\pm$3.8 & 18, 21, 11, 25, 29 \\
H1 & Full context & 96.9 & 96.7 & 96.5 & 96.9 & 95.1 & 96.4$\pm$0.7 & 16, 10, 11, 12, 10 \\
\addlinespace
H2 & SALM & 89.8 & 84.4 & 90.4 & 93.8 & 89.5 & 89.6$\pm$3.4 & 30, 19, 24, 30, 20 \\
H2 & Full context & 93.6 & 93.6 & 93.8 & 93.9 & 94.3 & 93.8$\pm$0.3 & 9, 23, 16, 14, 12 \\
\addlinespace
H3 & SALM & 91.2 & 88.5 & 86.5 & 95.3 & 94.3 & 91.2$\pm$3.7 & 16, 25, 22, 28, 24 \\
H3 & Full context & 95.7 & 92.6 & 96.3 & 96.1 & 95.3 & 95.2$\pm$1.5 & 22, 5, 13, 14, 12 \\
\addlinespace
H4 & SALM & 97.7 & 95.5 & 98.4 & 97.9 & 96.5 & 97.2$\pm$1.2 & 23, 16, 18, 22, 20 \\
H4 & Full context & 96.7 & 95.9 & 97.5 & 96.3 & 95.9 & 96.4$\pm$0.7 & 13, 12, 15, 11, 15 \\
\addlinespace
H5 & SALM & 74.2 & 71.1 & 76.8 & 70.9 & 73.8 & 73.4$\pm$2.4 & 27, 26, 29, 22, 28 \\
H5 & Full context & 94.9 & 94.3 & 93.6 & 93.9 & 94.3 & 94.2$\pm$0.5 & 11, 8, 12, 10, 9 \\
\bottomrule
\end{tabular}

\end{table}

\subsection{Write interfaces}
\label{app:adapter_details}

Table~\ref{tab:params} gives the trainable parameters of each arm by component, read from the optimizer
parameter groups of each run; the frozen GPT-2 (124M parameters) is excluded. The read encoder (attention
pooling over 768-dimensional token states) and the GRU transition ($768\to64$) are identical across SALM
arms. FiLM uses a generator of width 981, prefix tuning eight key/value prefix tokens per layer through a
width-128 generator, and deep prompting eight prompt tokens per layer through a width-256 generator; each
setting is the single integer capacity that best matches generated LoRA's total. Cross-attention uses a
768-wide attention module and is not parameter-matched.

\begin{table}[h]
\caption{Trainable parameters by component (from each run's optimizer parameter groups).}
\label{tab:params}
\centering\small
\begin{tabular}{lrrrr}
\toprule
Arm & Encoder & Transition & Write / LoRA & Total \\
\midrule
Generated LoRA & 591{,}360 & 160{,}128 & 1{,}604{,}032 & 2{,}355{,}520 \\
Gated LoRA & 591{,}360 & 160{,}128 & 1{,}618{,}692 & 2{,}370{,}180 \\
Soft prompt & 591{,}360 & 160{,}128 & 1{,}602{,}942 & 2{,}354{,}430 \\
FiLM & 591{,}360 & 160{,}128 & 1{,}603{,}893 & 2{,}355{,}381 \\
Prefix tuning & 591{,}360 & 160{,}128 & 1{,}597{,}952 & 2{,}349{,}440 \\
Deep prompt & 591{,}360 & 160{,}128 & 1{,}604{,}224 & 2{,}355{,}712 \\
Cross-attention & 591{,}360 & 160{,}128 & 2{,}571{,}712 & 3{,}323{,}200 \\
Full-context LoRA & 0 & 0 & 2{,}359{,}296 & 2{,}359{,}296 \\
\bottomrule
\end{tabular}

\end{table}

\subsection{Longer episodes without retraining}
\label{app:long_horizon}

Four-turn HCM models were evaluated, without retraining or length-specific selection, on episodes of
$n\in\{4,6,8,12\}$ turns ($n-1$ hard truthful clues and the read-only query). Code pool, evidence
templates and answer format are those of training; only the generator seed differs, and no evaluation
evidence sequence occurs in the training data. Condition~A applies the training acceptance rule to the
longer evidence, so most added clues are redundant. Condition~B additionally requires that the first clue
be necessary: without it at least two codes remain. All models of a condition and length score the same
\V{lh/n} episodes. Table~\ref{tab:long_horizon} reports seeds 0--2 of three SALM interfaces and the
baseline; the ten-seed condition-B comparison of Figure~\ref{fig:long_horizon} uses the same episodes for
seeds 0--2 (its per-seed values reproduce those in the table). Under condition~B the gap between gated
LoRA and full-context LoRA is \V{lh/B4/diff} points at four turns (\V{lh/B4/diffci}, ten seeds),
\V{lh/B6/diff} at six (\V{lh/B6/diffci}) and \V{lh/B8/diff} at eight (\V{lh/B8/diffci}). Zeroing the
state at the query reduces gated LoRA to \V{lh/zero/min}--\V{lh/zero/max}\% at these lengths, so
the longer-episode predictions still depend on the state.

\begin{table}[h]
\caption{Accuracy (\%) of four-turn-trained models on longer episodes, mean $\pm$ s.d.\ over seeds 0--2.
A: redundant added evidence. B: the first clue is necessary.}
\label{tab:long_horizon}
\centering\small
\begin{tabular}{llcccc}
\toprule
Cond. & Arm & $n{=}4$ & $n{=}6$ & $n{=}8$ & $n{=}12$ \\
\midrule
A & Generated LoRA & 81.0$\pm$5.5 & 78.4$\pm$6.3 & 79.9$\pm$5.5 & 79.9$\pm$3.6 \\
A & Gated LoRA & 86.2$\pm$2.5 & 84.6$\pm$1.8 & 83.1$\pm$3.7 & 78.9$\pm$4.3 \\
A & Soft prompt & 84.6$\pm$4.0 & 79.9$\pm$1.6 & 80.2$\pm$2.0 & 79.2$\pm$2.3 \\
A & Full-context LoRA & 89.8$\pm$1.4 & 90.1$\pm$1.2 & 86.5$\pm$0.9 & 83.6$\pm$2.3 \\
\midrule
B & Generated LoRA & 76.3$\pm$4.6 & 60.2$\pm$5.5 & 61.2$\pm$3.5 & 56.5$\pm$2.4 \\
B & Gated LoRA & 81.2$\pm$3.4 & 64.6$\pm$6.0 & 61.5$\pm$3.6 & 55.7$\pm$4.3 \\
B & Soft prompt & 74.0$\pm$3.9 & 58.3$\pm$7.1 & 56.2$\pm$2.7 & 54.9$\pm$4.3 \\
B & Full-context LoRA & 88.8$\pm$0.9 & 87.0$\pm$0.5 & 87.2$\pm$1.2 & 84.6$\pm$11.1 \\
\bottomrule
\end{tabular}

\end{table}

\subsection{Inference cost}
\label{app:compute}

We timed the seed-0 checkpoints of H1--H3 ($M=16$; $T=4,8,16$) on CPU (float32, batch 1, four threads,
32 test episodes, one process per measurement, model loading excluded). Table~\ref{tab:compute} reports
median end-to-end latency per episode, SALM's median cost per assimilation step, the state that SALM
carries between observations, and the key--value cache implied by the baseline's transcript. SALM is
\V{cmp/ratiomin}--\V{cmp/ratiomax}$\times$ slower at these lengths, because every observation is encoded
by a full GPT-2 pass and every prediction applies a generated adaptation. Its persistent state is
\V{cmp/statekb}~KB at every $T$, against \V{cmp/kvmin}--\V{cmp/kvmax}~MB of key--value cache for the
transcript. Peak process memory was \V{cmp/peakmin}--\V{cmp/peakmax}~GB for both models, dominated by the shared frozen
backbone. Longer horizons, batching and GPU throughput were not measured.

\begin{table}[h]
\caption{CPU inference cost, seed-0 checkpoints, batch 1.}
\label{tab:compute}
\centering\small
\begin{tabular}{lrrrrr}
\toprule
Cell ($T$) & SALM latency (ms) & FC latency (ms) & SALM step (ms) & SALM state (KB) & FC KV cache (MB) \\
\midrule
H1 (4) & 640 & 78 & 138 & 3.3 & 3.1 \\
H2 (8) & 768 & 121 & 90 & 3.3 & 6.0 \\
H3 (16) & 1551 & 228 & 93 & 3.3 & 12.1 \\
\bottomrule
\end{tabular}

\end{table}

\section{Stochastic Experiment: Details and Additional Results}
\label{app:sto_details}

\paragraph{Protocol.} The protocol was fixed before training. Splits: 512 training and 128 validation
episodes, both excluding 20 unordered answer patterns that were held out at random; a natural test split of
\V{sto/n} episodes; an unseen-pattern test split of \V{sto/nu} episodes containing only the held-out
patterns. Each run keeps the epoch with the lowest validation oracle KL of the mixture at the training
particle count, and stops after six epochs without an improvement of 0.002 nats (minimum 10, maximum 60
epochs). Optimization: AdamW, learning rate $3\times10^{-4}$, weight decay 0.01, batch 4, gradient clipping 1.
Evaluation uses fixed Monte Carlo seeds (common random numbers across models). The transition scale has
minimum 0.01; the four mixture components of the initial law are fixed. Seeds 3 and 4 were trained later
with the original configuration files; every split fingerprint was checked against the original manifest
before training.

\paragraph{Convergence.} All runs stopped on patience. The stochastic runs ran for \V{sto/sto/epochs}
epochs and selected epochs \V{sto/sto/best} (seeds 0--4); the deterministic runs ran for \V{sto/det/epochs}
epochs and selected \V{sto/det/best}. Mean CPU training time over seeds 0--2 was \V{sto/sto/hours}~h
(stochastic) and \V{sto/det/hours}~h (deterministic).

\paragraph{Particle count at evaluation.} Table~\ref{tab:sto_K} evaluates the trained models with $K=1$,
4 and 8 particles. A single particle is a confident commitment to one code and is far worse than uniform;
four and eight particles are nearly equivalent, as expected when the four mixture components already cover
the four hypotheses.

\begin{table}[h]
\caption{Oracle KL (nats) on the natural test split by number of particles at evaluation.}
\label{tab:sto_K}
\centering\small
\begin{tabular}{lccc}
\toprule
Variant (seeds) & $K{=}1$ & $K{=}4$ (training) & $K{=}8$ \\
\midrule
Stochastic, oracle CE (5) & 5.333$\pm$0.838 & 0.009$\pm$0.002 & 0.008$\pm$0.002 \\
Stochastic, oracle CE + transport (3) & 2.368$\pm$0.146 & 0.010$\pm$0.002 & 0.008$\pm$0.002 \\
Deterministic, oracle CE (5) & 0.042$\pm$0.007 & -- & -- \\
\bottomrule
\end{tabular}

\end{table}

\paragraph{Transport ablation.} A third variant adds $0.25\,\mathcal L_{\mathrm{transport}}$ to
Equation~\eqref{eq:stochastic_objective}: the cost $\sum_{k,c}\pi_{kc}[-\log q_k(c)]$ of an optimal coupling
$\pi$ between the particle weights and the exact posterior, with the coupling held fixed in the gradient. It
was trained with seeds 0--2 only. Its test oracle KL is \V{sto/tr/kl} nats, against \V{sto/sto/kl} without
transport (paired difference over seeds 0--2: \V{sto/diff/trkl}, \V{sto/diff/trklci}). Particle
specialization therefore does not depend on the transport term; without it a single particle is, if
anything, more committed (Table~\ref{tab:sto_K}).

\paragraph{Interventions.} Replacing the state at the query by the initial state raises oracle KL to
\V{sto/sto/zero_kl} (stochastic) and \V{sto/det/zero_kl} (deterministic); replacing it by another episode's
state raises it to \V{sto/sto/shuf_kl} and \V{sto/det/shuf_kl}. Both models depend on their state.

\paragraph{Unseen answer patterns.} On the unseen-pattern split the exact posterior reaches accuracy
\V{sto/bayes/accu}\% and NLL \V{sto/bayes/nllu}. The stochastic model reaches oracle KL \V{sto/sto/klu},
NLL \V{sto/sto/nllu} and accuracy \V{sto/sto/accu}\%; the deterministic model \V{sto/det/klu},
\V{sto/det/nllu} and \V{sto/det/accu}\%. Both select a posterior mode in nearly every episode
(\V{sto/sto/mapagreeu}\% and \V{sto/det/mapagreeu}\%), and \V{sto/tiedu}\% of these episodes have tied modes.

\paragraph{Scope of the stochastic results.} The comparison is limited to $M=K=4$ with one particle per
mixture component, a single reliability $\rho=0.8$ and three observations. Larger hypothesis spaces, in which
particles must represent several hypotheses each, and training without the exact posterior are under
investigation. Preliminary oracle-free results, including eight hypotheses with four particles, are reported
separately in Appendix~\ref{app:oracle_free}.

\section{Oracle-Free Stochastic SALM (Preliminary)}
\label{app:oracle_free}

Both models in Section~\ref{sec:stochastic} are trained with the exact posterior as an auxiliary target. This
appendix reports an earlier, separately run study in which the stochastic SALM is trained \emph{without} it.
The results are preliminary: three seeds per arm, one test split and one task family, with an evidence
format that differs from Section~\ref{sec:stochastic}. They are not a controlled replication of that section,
and the two sets of numbers should not be compared directly.

\paragraph{Task.} The four codes $\{44,19,93,90\}$, the three observations, the reliability $\rho=0.8$ and the
exact posterior of Equation~\eqref{eq:noisy_posterior} are as in Section~\ref{sec:stochastic}. Each
observation is a natural-language report by a narrator, for example \emph{Alice reports that the statement
``the codeword is an even number'' is false.} Here $x_t$ is the statement up to its last word and $y_t$ is the
truth word (\emph{true} or \emph{false}), which is correct for the hidden code with probability $\rho$. The
predicates are parity and thresholds between codes, with paraphrased wording; no test statement occurs
verbatim in the training data. The splits have 512 training, 128 validation and 512 test episodes, with no
held-out answer patterns. In \V{of/tied4}\% of test episodes two or more codes share the maximal exact
posterior probability.

\paragraph{Model and training.} The stochastic model is that of Section~\ref{sec:stochastic}: a 16-dimensional
state, the residual Gaussian transition, a fixed four-component initial mixture with particles drawn one per
component, the bootstrap particle filter without resampling, and a gated-LoRA write. Training uses AdamW
(learning rate $3\times10^{-4}$, weight decay 0.01, batch 8) and stops after six epochs without a 0.002-nat
validation improvement (minimum 10, maximum 40 epochs). Each run keeps the epoch with the lowest
\emph{validation NLL of the sampled code}. Unlike Section~\ref{sec:stochastic}, checkpoint selection never uses
the exact posterior, in any arm. All runs stopped on patience.

\paragraph{Objectives.} The three arms use the first one, two or three terms of
Equation~\eqref{eq:stochastic_objective}:
\begin{itemize}
\item \textbf{A}, oracle-free: $\mathcal L_{\mathrm{recall}}$ only, the NLL of the sampled code.
\item \textbf{B}, oracle-free: $\mathcal L_{\mathrm{recall}}+\mathcal L_{\mathrm{evidence}}$, adding the
filtering-variational likelihood of the observed truth words.
\item \textbf{C}, oracle-supervised: $\mathcal L_{\mathrm{recall}}+\mathcal L_{\mathrm{evidence}}+\mathcal
L_{\mathrm{oracle}}$, adding the cross-entropy to the exact posterior.
\end{itemize}
In A and B the exact posterior is computed only at evaluation. The logged loss terms of every optimizer step
confirm that no oracle term entered training in either arm.

Objective B was used for the two scaling ablations. It was chosen before any test evaluation by a rule fixed in
advance that compares A and B on validation data; C was not eligible. The rule's primary criterion,
validation NLL, did not separate A and B within seed noise, so the choice fell to its secondary criterion,
\emph{validation} oracle KL. The exact posterior therefore informed which objective the ablations use, but it
did not enter the training or checkpoint selection of any ablation model.

The two oracle-free ablations change one factor each:
\begin{itemize}
\item $(M,K)=(8,4)$: eight codes, four particles and six observations, on its own data. Hypotheses outnumber
particles.
\item $(M,K)=(4,8)$: the four-code test set of the main arms, with eight particles drawn two per component of
the four-component initial mixture.
\end{itemize}

\paragraph{Metrics.} The metrics are those of Section~\ref{sec:stochastic}, plus four more.
\begin{itemize}
\item The \emph{effective sample size} $1/\sum_k(w^{(k)})^2$ of the final weights.
\item The number of \emph{distinct particle MAPs}: the number of different codes among the $K$ particles' top
codes $\arg\max_cq_k(c)$, at most $\min(K,M)=4$.
\item \emph{Codes per particle}: the effective number of codes $\exp H(q_k)$ in a particle's own prediction.
\item The \emph{shuffle gap}: the increase in NLL when the state at the query is replaced by another
episode's, read from the run records.
\end{itemize}
All metrics except the shuffle gap were recomputed from the stored per-episode predictions.

\begin{table}[h]
\caption{Oracle-free stochastic SALM (preliminary). Test split of 512 episodes; mean $\pm$ s.d.\ over three
seeds. \emph{Oracle in loss}: whether $\mathcal L_{\mathrm{oracle}}$ was used in training; checkpoints are
always selected on validation NLL of the sampled code. Distinct particle MAPs are at most 4, and codes per
particle at most $M$. Exact-posterior and uniform rows are references for each task.}
\label{tab:oracle_free}
\centering\small
\setlength{\tabcolsep}{2.6pt}
\begin{tabular}{llcccccccc}
\toprule
Task & Objective & Oracle & Accuracy (\%) & Oracle KL & NLL & Shuffle gap & ESS & Distinct & Codes per \\
$(M, K)$ & & in loss & $\uparrow$ & $\downarrow$ & $\downarrow$ & (nats) & & particle MAPs & particle \\
\midrule
(4, 4) & A: $\mathcal{L}_{\mathrm{recall}}$ & no & 45.6$\pm$5.8 & 0.246$\pm$0.056 & 1.150$\pm$0.054 & 0.75$\pm$0.15 & 1.90$\pm$0.12 & 3.34$\pm$0.36 & 2.18$\pm$0.29 \\
(4, 4) & B: A $+\,\mathcal{L}_{\mathrm{evidence}}$ & no & 49.0$\pm$1.4 & 0.197$\pm$0.021 & 1.109$\pm$0.018 & 0.88$\pm$0.01 & 2.01$\pm$0.02 & 3.82$\pm$0.16 & 1.80$\pm$0.02 \\
(4, 4) & C: B $+\,\mathcal{L}_{\mathrm{oracle}}$ & yes & 52.2$\pm$1.0 & 0.108$\pm$0.013 & 1.011$\pm$0.019 & 0.99$\pm$0.10 & 2.14$\pm$0.08 & 3.89$\pm$0.12 & 1.67$\pm$0.10 \\
\midrule
(8, 4) & B (selected) & no & 27.0$\pm$0.3 & 0.383$\pm$0.067 & 1.766$\pm$0.046 & 0.91$\pm$0.09 & 1.87$\pm$0.10 & 3.76$\pm$0.33 & 3.87$\pm$0.42 \\
(4, 8) & B (selected) & no & 47.9$\pm$2.2 & 0.185$\pm$0.026 & 1.096$\pm$0.026 & 0.87$\pm$0.02 & 4.13$\pm$0.04 & 3.55$\pm$0.35 & 1.79$\pm$0.04 \\
\midrule
$M{=}4$ & Exact posterior & -- & 57.8 & 0 & 0.904 & -- & -- & -- & -- \\
$M{=}4$ & Uniform & -- & 25.0 & 0.469 & 1.386 & -- & -- & -- & -- \\
$M{=}8$ & Exact posterior & -- & 46.9 & 0 & 1.380 & -- & -- & -- & -- \\
$M{=}8$ & Uniform & -- & 12.5 & 0.684 & 2.079 & -- & -- & -- & -- \\
\bottomrule
\end{tabular}

\end{table}

\paragraph{Learning without the oracle.} Table~\ref{tab:oracle_free} reports the results. Both oracle-free
objectives learn the task well above chance and depend on their state.
\begin{itemize}
\item \textbf{Accuracy.} A reaches \V{of/A/acc}\% and B \V{of/B/acc}\%, against \V{of/bayes4/acc}\% for the
exact posterior and 25\% for uniform. As in Section~\ref{sec:stochastic}, accuracy against the sampled code
partly reflects how tied codes are broken.
\item \textbf{Fit to the exact posterior.} Oracle KL is \V{of/A/kl} (A) and \V{of/B/kl} (B), removing
\V{of/A/fracremoved}\% and \V{of/B/fracremoved}\% of the uniform prediction's divergence (\V{of/unif4/kl}
nats).
\item \textbf{State dependence.} Replacing the state by another episode's raises NLL by \V{of/A/shuffle_gap}
and \V{of/B/shuffle_gap} nats.
\item \textbf{A against B.} The difference is not resolved with three seeds: B $-$ A oracle KL is
\V{of/diff/B-A/kl}, Welch 95\% interval \V{of/diff/B-A/klci}. The spread of A is large; its seeds range from
\V{of/A/accmin}\% to \V{of/A/accmax}\% accuracy.
\item \textbf{With the oracle term.} C lowers oracle KL to \V{of/C/kl} (\V{of/C/fracremoved}\% removed; C $-$ B
\V{of/diff/C-B/kl}, \V{of/diff/C-B/klci}), and every C seed is below every B seed. Posterior supervision still
accounts for much of the fit; the oracle-free models are far from exact posterior inference.
\end{itemize}

\paragraph{Distinct particle predictions persist without oracle supervision.} In every oracle-free run the
particles keep different predictions.
\begin{itemize}
\item \textbf{Distinct top codes.} The four particles predict on average \V{of/A/distinct_particle_maps} (A)
and \V{of/B/distinct_particle_maps} (B) distinct top codes per test episode, out of four, against
\V{of/C/distinct_particle_maps} with the oracle term. No seed falls below \V{of/A/distinct_particle_mapsmin}.
\item \textbf{Weights.} The final effective sample size is about two of four (\V{of/A/ess} and \V{of/B/ess}),
so after three observations the weights concentrate, on average, on about two particles.
\item \textbf{Specialization.} The particles are less specialized than in Section~\ref{sec:stochastic}. Each
spreads over \V{of/A/particle_effective_codes} (A) and \V{of/B/particle_effective_codes} (B) effective codes,
with mean top probability \V{of/A/particle_max_prob} and \V{of/B/particle_max_prob}, against
\V{of/C/particle_effective_codes} and \V{of/C/particle_max_prob} for C.
\end{itemize}
A mixture of particles with distinct predictions, reweighted by the evidence, therefore emerges from the recall
objective alone; it does not require a posterior target. One design choice may contribute: the stratified
initial law places the particles in different mixture components from the start. All runs here use it, so
its effect is not separated.

\paragraph{Scaling ablations (oracle-free, objective B).}
\begin{itemize}
\item \textbf{Eight codes, four particles.} The fit is much weaker. Accuracy is \V{of/M8K4/acc}\%, against
\V{of/bayes8/acc}\% for the exact posterior and 12.5\% for uniform. Oracle KL is \V{of/M8K4/kl}, removing only
\V{of/M8K4/fracremoved}\% of the uniform divergence (\V{of/unif8/kl}). The particles remain distinct, with
\V{of/M8K4/distinct_particle_maps} distinct top codes out of four, but each spreads over
\V{of/M8K4/particle_effective_codes} of the eight codes. Particles thus cover overlapping sets of hypotheses
rather than one code each. This task has more codes and more observations than the four-code task, so its
numbers are not directly comparable.
\item \textbf{Four codes, eight particles.} Doubling the particles on the four-code task gives oracle KL
\V{of/M4K8/kl}, against \V{of/B/kl} with four (difference \V{of/diff/M4K8-B/kl}, \V{of/diff/M4K8-B/klci}).
Accuracy is \V{of/M4K8/acc}\%, effective sample size \V{of/M4K8/ess} of eight, and
\V{of/M4K8/distinct_particle_maps} distinct top codes. The extra particles did not measurably change the fit.
\end{itemize}

\paragraph{Deterministic reference.} A deterministic SALM ($K=1$) trained with objective B on this protocol
stayed at chance: accuracy \V{of/det/acc}\%, oracle KL \V{of/det/kl} (uniform \V{of/unif4/kl}), and a shuffle
gap of \V{of/det/shuffle_gap} nats in every seed (selected epochs \V{of/det/best}). We did not diagnose this
failure. The protocol therefore provides no stochastic-versus-deterministic comparison, and we draw none from
it.

\paragraph{Convergence.} Training ran for the following numbers of epochs per seed, with the selected epoch
in parentheses:
\begin{itemize}
\item A: \V{of/A/epochs} (\V{of/A/best});
\item B: \V{of/B/epochs} (\V{of/B/best});
\item C: \V{of/C/epochs} (\V{of/C/best});
\item $(8,4)$: \V{of/M8K4/epochs} (\V{of/M8K4/best});
\item $(4,8)$: \V{of/M4K8/epochs} (\V{of/M4K8/best}).
\end{itemize}

\paragraph{Interpretation and limitations.} Without any posterior supervision, the particle model learns
evidence-dependent predictions that remove roughly half of the uniform prediction's divergence from the exact
posterior. It also keeps distinct particle hypotheses that the weights select among. The qualitative structure
of Section~\ref{sec:stochastic}, a reweighted bank of distinct hypotheses, thus does not depend on the oracle
target. Calibrated posterior inference without supervision is not established, for four reasons:
\begin{itemize}
\item the oracle-supervised arm fits the posterior substantially better;
\item the fit degrades when hypotheses outnumber particles;
\item all comparisons rest on three seeds;
\item the evidence format, selection criterion and data differ from Section~\ref{sec:stochastic}.
\end{itemize}

\section{Preliminary Natural-Language Experiment: MultiRC}
\label{app:multirc}

\paragraph{Status.} This experiment is preliminary. It is included for transparency and is not used to
support any claim of the paper.

\paragraph{Setup.} MultiRC \citep{khashabi2018multirc,wang2019superglue} pairs passages with questions that
have several candidate answers, each labelled correct or incorrect. For each candidate-answer episode the
passage is split into three contiguous, sentence-aligned fragments $D_1,D_2,D_3$ of balanced GPT-2 length,
and a fourth, read-only query asks whether the candidate answers the question. SALM updates its
64-dimensional state from the fragments, and its decoder sees only the question and candidate; full-context
LoRA prepends the same fragment tokens to the query. The scored outcomes are the tokens \texttt{yes} and
\texttt{no}, renormalized over these two outcomes, and only this answer is supervised. Both arms use the frozen
GPT-2 small backbone, gated LoRA (rank 43, $\alpha=86$) for SALM and static LoRA (rank 64, $\alpha=128$) for the
baseline, and are trained from scratch on 2{,}052 training pairs drawn from a passage-disjoint training pool,
with checkpoint selection on 259 internal-validation pairs (minimum binary NLL; AdamW, learning rate
$10^{-4}$, weight decay 0.01, batch 8, patience 4, 6--40 epochs, float32 on T4 GPUs). The labelled official validation split (\V{mrc/npairs} pairs, \V{mrc/npassages} passages) was held
out from training and selection and is reported as \emph{held-out dev.} We report \V{mrc/nseeds} matched model
seeds on one fixed passage partition.

\begin{table}[h]
\caption{MultiRC, mean $\pm$ s.d.\ over \V{mrc/nseeds} seeds. NLL is binary, in nats; accuracy and answer-level
F1 (F1a) are over candidate answers; EM is the fraction of questions whose candidates are all correct.}
\label{tab:multirc}
\centering\small
\begin{tabular}{llcccc}
\toprule
Model & Split & NLL & Accuracy (\%) & F1a (\%) & EM (\%) \\
\midrule
SALM (Gated LoRA) & Internal val. & 0.628$\pm$0.009 & 65.3$\pm$2.4 & 57.3$\pm$7.2 & 8.9$\pm$4.3 \\
SALM (Gated LoRA) & Held-out dev. & 0.642$\pm$0.007 & 63.5$\pm$1.4 & 53.2$\pm$3.7 & 8.0$\pm$1.6 \\
Full-context LoRA & Internal val. & 0.571$\pm$0.021 & 70.4$\pm$2.0 & 65.0$\pm$7.1 & 16.1$\pm$5.2 \\
Full-context LoRA & Held-out dev. & 0.592$\pm$0.007 & 66.9$\pm$1.6 & 61.6$\pm$6.2 & 12.5$\pm$2.7 \\
\bottomrule
\end{tabular}

\end{table}

\paragraph{Results.} On held-out pairs SALM obtains NLL \V{mrc/salm/nll} and F1a \V{mrc/salm/f1}\%, against
\V{mrc/fc/nll} and \V{mrc/fc/f1}\% for full-context LoRA (Table~\ref{tab:multirc}); the paired NLL difference is
\V{mrc/diff} nats (95\% hierarchical bootstrap over seeds and passages \V{mrc/diffci}). Zeroing SALM's state
raises held-out NLL to \V{mrc/salm/zeronll}, so the predictions depend on having a state. Replacing the state by
that of another passage changes NLL by only \V{mrc/shuf} nats (\V{mrc/shufci}). For the baseline, removing the
passage raises NLL to \V{mrc/fc/nopassnll} and shuffling passages to \V{mrc/fc/shufnll}.

\paragraph{Interpretation and limitations.} SALM learns a useful state-conditioned predictor on this task
but uses little passage-specific information from its state; the difference between the zero-state and
shuffled-state effects indicates that much of the state's influence is shared across passages. We have run
further diagnostics and ablations of the encoder, transition and adapter, but the mechanism behind the weak
passage-specific dependence has not been established, and we do not speculate on it here. The results cover
one partition, four seeds and a small training set.

\end{document}